\documentclass[acmtog,nonacm]{acmart}

\usepackage{bm}
\usepackage{pifont}
\usepackage{algorithm}
\usepackage{algorithmic}

\definecolor{keyframepink}{cmyk}{0.60, 0.90, 0, 0}
\definecolor{burstblue}{cmyk}{1, 0.90, 0, 0}
\newcommand{\newtext}[1]{{#1}}
\begin{document}
\title{Keyframe-Centric State-Space Modeling for Burst Image Super-Resolution}

\author{Ozan Unal}
\email{ozan.unal@huawei.com}
\affiliation{%
  \department{Computer Vision Lab}
  \institution{Huawei}
  \city{Zurich}
  \country{Switzerland}
}

\author{Steven Marty}
\email{steven.marty@h-partners.com}
\affiliation{%
  \department{Computer Vision Lab}
  \institution{Huawei}
  \city{Zurich}
  \country{Switzerland}
}

\author{Dengxin Dai}
\email{dengxin.dai@huawei.com}
\affiliation{%
  \department{Computer Vision Lab}
  \institution{Huawei}
  \city{Zurich}
  \country{Switzerland}
}
\authorsaddresses{}

\begin{teaserfigure}
  \centering
  \includegraphics[width=.99\textwidth]{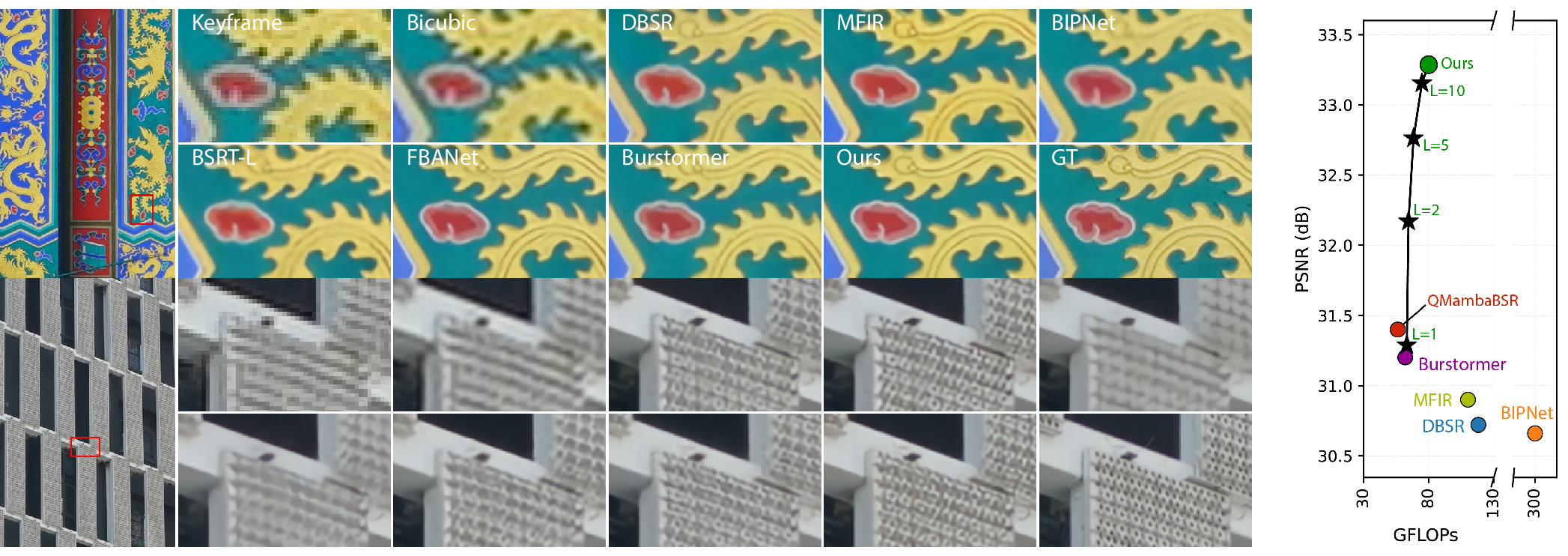}
  \caption{BurstMamba decouples keyframe super-resolution from burst prior extraction: a high-capacity Mamba-based SISR stream reconstructs the HR keyframe, while a lightweight burst stream routes only sub-pixel evidence from additional frames and injects it as a residual refinement. Left: On RealBSR-RGB ($\times4$), BurstMamba recovers fine structures and high-frequency detail (thin lines, small gaps) that are missed or oversmoothed by prior burst methods. Right: PSNR-GFLOPs trade-off under a common protocol (14 frames \newtext{unless stated otherwise}). BurstMamba scales favorably with burst length $L$, improving monotonically from $L{=}1$ to $L{=}14$, while adding only 1.3 GFLOPs per extra frame.}
  \Description{Comparison of BurstMamba against other BISR methods: qualitatively on RealBSR-RGB and computationally on a PSNR-GFLOPs trade-off graph.}
  \label{fig:teaser}
\end{teaserfigure}

\begin{abstract}
  Burst image super-resolution (BISR) reconstructs a high-resolution keyframe by aggregating complementary sub-pixel evidence from a short burst of low-resolution frames. Existing methods often process all burst frames with heavy backbones or maintain deep cross-frame interaction throughout the network, leading to redundant computation on non-key frames and limiting scalability with burst length. In this work we propose BurstMamba, a BISR architecture built on a simple principle: allocate most compute to reconstructing the keyframe, and process the burst primarily to extract sub-pixel priors. To this end, BurstMamba decouples BISR into a high-capacity keyframe super-resolution stream and a lightweight burst stream that interacts with it only through stage-wise residual injection. To improve burst-to-keyframe transfer, we introduce Gather $\rightarrow$ Aggregate $\rightarrow$ Scatter (GAS), which uses correspondence only for cross-frame message passing while preserving native-view features through a residual connection, and a wavelet-conditioned state update that biases selective routing toward high-frequency regions. Across SyntheticSR, RealBSR-RAW, and RealBSR-RGB, BurstMamba achieves state-of-the-art results. Extensive ablations show that the proposed compute separation, GAS, and wavelet-conditioned routing each contribute to the model's accuracy and scaling.
\end{abstract}

\maketitle

\section{Introduction}

Burst image super-resolution (BISR) reconstructs a single high-resolution (HR) \emph{keyframe} by leveraging a short burst of low-resolution (LR) frames captured in rapid succession~\cite{wronski2019handheld,lafenetre2023handheld,valsesia2021permutation}. Compared to single-image super-resolution (SISR), the burst provides multiple observations of the same scene under slight viewpoint jitter, noise realizations, and local motion, supplying complementary \emph{sub-pixel evidence} that reduces ambiguity and enables sharper recovery of high-frequency texture, especially under aggressive upscaling and challenging fine detail\newtext{~\cite{bhaticcv2023}}.

Recent progress in BISR has been primarily driven by architectural improvements in the way burst information is aggregated. Early learning-based approaches used convolutional alignment-and-fusion pipelines~\cite{luo2021ebsr,mehta2022adaptive,dudhane2022burst}, whereas current state-of-the-art methods increasingly rely on transformer backbones to model long-range interactions across space and time~\cite{luo2022bsrt,dudhane2023burstormer}. While attention is expressive, its quadratic scaling in the number of tokens makes high-resolution restoration expensive, and practical variants often introduce windowing or sparsification~\cite{liu2021swin,tay2022efficient} that may trade global context, numerical stability, or accuracy for efficiency~\cite{wang2020linformer}.

Selective state-space models (SSMs) offer an appealing alternative: by using input-dependent state updates, Mamba (S6) enables linear-time sequence modeling~\cite{gu2023mamba}, and with appropriate 2D scanning strategies, SSMs can match or surpass attention in image restoration at lower compute~\cite{liu2025vmamba,guo2024mambair,zhu2024visionmamba}. However, adopting efficient sequence models alone does not resolve a BISR-specific asymmetry that is often underemphasized. In BISR, the goal is to reconstruct \emph{one} high-quality HR keyframe; the remaining frames contribute primarily as \emph{priors} that provide complementary sub-pixel evidence. This distinction matters for compute allocation: much of the costly spatial reconstruction should be reserved for the keyframe, whereas non-key frames need only be processed to the extent required to extract and transfer useful priors. Yet many existing designs either process every burst frame with a deep backbone before fusion~\cite{bhat2021deep,wei2023towards} or maintain heavy cross-frame interaction throughout the network~\cite{dudhane2023burstormer}, leading to redundant computation and limiting scalability with burst length.

This asymmetry suggests a simple principle: allocate the expensive spatial reconstruction to the keyframe, and treat the remaining frames as a source of compact, transferable priors. We therefore elevate this \textit{decoupling} to an explicit design axis and propose BurstMamba, a transformer-free BISR architecture\newtext{, leveraging the strong linear routing of a Mamba backbone}, that separates keyframe reconstruction from burst prior extraction . BurstMamba comprises two streams: (1) a high-capacity SISR backbone, and (2) a lightweight burst module that processes the burst to distill per-pixel residual priors. At each stage, the burst stream refines its internal representation and produces an updated prior for the keyframe, while the keyframe stream continues to improve the reconstruction. This interaction is strictly one-way, with information flowing only from the burst to the keyframe. This design enforces a clear division of labor: the keyframe stream focuses on high-quality reconstruction, while the burst stream specializes in extracting and routing useful information from the remaining frames.

Two designs increase the effectiveness of this decoupling in practice. First, we introduce \textit{Gather $\rightarrow$ Aggregate $\rightarrow$ Scatter (GAS).} Alignment is a standard practice in BISR\cite{bhat2021deep,wei2023towards, di2024qmambabsrburstimagesuperresolution}. However, explicitly warping frames/features upfront can smooth away sub-pixel cues. GAS changes \emph{where} alignment is applied: each frame is processed in its native viewpoint, and correspondence is used only during cross-frame aggregation. For each keyframe pixel, we gather features from all frames at corresponding (sub-pixel) locations using bilinear sampling, aggregate them across the burst, and scatter the result back to each frame's native coordinates as \textit{residuals}, ensuring the preservation of native viewpoint details. This avoids early warping while still enabling accurate cross-frame information exchange. Second, we propose a wavelet-conditioned state update ($\psi$S6) that biases cross-frame aggregation toward high-frequency evidence. We compute a one-level Dual-Tree Complex Wavelet Transform (DTCWT) from burst features to parameterize Mamba's selectivity, prioritizing burst-to-keyframe transfer in textured regions by limiting model deepening.

At a high level, BurstMamba is intentionally simple: it replaces implicit, entangled burst processing with an explicit factorization and a minimal interface. \newtext{As with Transformers in prior restoration work, the operator alone is not our contribution. Mamba nevertheless plays a crucial role in BurstMamba: it provides the efficient routing operator responsible for linear scaling with burst length and serves as the foundation for our wavelet-based reformulation. Our central contribution is the set of constraints imposed on how burst computation is allocated and routed relative to the keyframe: (i) reconstruction compute is restricted to the keyframe, (ii) burst information is routed through structured per-pixel state updates, and (iii) alignment is used only where indexing consistency is required.} These constraints yield properties that do not emerge from standard fusion blocks, including linear scaling in burst length and the ability to detach the burst module for SISR fallback.

BurstMamba achieves state-of-the-art results on public BISR benchmarks: RealBSR-RGB~\cite{wei2023towards}, RealBSR-RAW~\cite{wei2023towards}, and SyntheticSR~\cite{bhat2021deep}. It scales favorably with burst length: achieving 31.29\,dB with $L{=}1$ (keyframe only) with the SR backbone having 20.6M parameters and costing 63 GFLOPs, and improving monotonically to 33.29\,dB as $L$ increases to 14, while compute and memory growing linearly with $1.3$ GFLOPs and 0.79M parameters per additional frame (see Fig.~\ref{fig:teaser}). This scaling behavior directly reflects the proposed decomposition.

\noindent In summary, we make three contributions:
\begin{enumerate}
  \item Keyframe-centric factorization for BISR: We introduce a compute separation formulation where the keyframe is reconstructed by a high capacity SR backbone, while the burst is processed only to distill keyframe priors via a one-way, keyframe only residual interface.
  \item Gather $\rightarrow$ Aggregate $\rightarrow$ Scatter (GAS) for burst processing:  We introduce GAS, a correspondence-aware aggregation scheme that uses alignment only during cross-frame aggregation, limiting warped information scattering to residuals in order to preserve native viewpoint details.
  \item Wavelet-conditioned selective dynamics: conditioning SSM selectivity on high-frequency wavelet responses to prioritize sub-pixel evidence transfer without excessive deepening.
\end{enumerate}

\section{Related Work}
\label{sec:related_work}

Our work lies at the intersection of (i) single-image super-resolution backbones, (ii) burst/multi-frame super-resolution architectures, and (iii) alignment and frequency-aware fusion. We focus this review on the design axes most relevant to our core claim: \emph{BISR should allocate most capacity to keyframe reconstruction and use the burst primarily to extract and route sub-pixel priors}.

\noindentparagraph{Single-image super-resolution} (SISR) provides the reconstruction engine that many multi-frame systems ultimately rely on. Early deep SISR models introduced end-to-end CNN mappings (e.g., SRCNN~\cite{dong2016srcnn}), followed by deeper residual designs~\cite{lim2017edsr,zhang2018rcan,ledig2017photo,wang2018esrgan} and transformer backbones that improved long-range context modeling, notably SwinIR~\cite{liang2021swinir,liu2021swin} and subsequent variants~\cite{chen2023activating,lu2022transformer,zhang2022efficient,Zamir_2022_CVPR}. However, attention scales quadratically with token count, motivating alternatives for high-resolution processing. Selective state-space models (SSMs) offer such an alternative. Mamba (S6)~\cite{gu2023mamba} enables linear-time sequence modeling via input-dependent state updates, and recent vision adaptations (e.g., 2D selective scanning) show strong performance across restoration tasks~\cite{liu2025vmamba,guo2024mambair,zhu2024visionmamba}. We build on this line by using a Mamba-based SISR backbone as the \emph{keyframe reconstruction stream} in BISR and introducing a dedicated burst prior stream that routes burst-derived evidence into that backbone.

\noindentparagraph{Burst image super-resolution} (BISR) leverages multiple LR images to reduce the ill-posedness of SISR~\cite{yang2019deep}. Classical multi-frame SR approaches relied on explicit motion models and known degradations to reduce artifacts and improve robustness~\cite{tsai1984multiframe,irani1991improving,peleg1987improving,bascle1996motion,elad1997restoration,elad2001fast}. Subsequent work relaxed strong assumptions by coupling SR with registration or deblurring under more realistic settings~\cite{he2007nonlinear,pickup2007overcoming,faramarzi2013unified}. Deep learning shifted BISR toward data-driven architectures that learn alignment, fusion, and reconstruction jointly. A key design axis is the extent to which the burst is processed deeply: (i) \emph{mid/late fusion} processes frames largely independently and aggregates later~\cite{bhat2021deep,wei2023towards}, while (ii) \emph{deep fusion} enables repeated interactions among burst features throughout the network~\cite{dudhane2023burstormer,mehta2023gated,luo2022bsrt}. Both strategies, however, share a structural property that we identify as a bottleneck: they allocate comparable compute to every frame in the burst, regardless of whether that frame is the keyframe being reconstructed. BurstMamba decouples these roles: the keyframe stream performs full high-capacity SR, while the burst stream is a shallow module whose sole function is to extract and route sub-pixel priors into the keyframe backbone.

\noindentparagraph{Alignment for burst fusion: pre-warping vs. native-view} A core challenge in BISR is handling motion while preserving sub-pixel evidence. Pre-alignment approaches such as DBSR~\cite{bhat2021deep}(optical flow at the bottleneck) and FBANet ~\cite{wei2023towards, di2024qmambabsrburstimagesuperresolution} (homography before Siamese processing) simplify fusion but risk blurring or attenuating high-frequency cues in the warping step. Several methods instead avoid explicit warping and support alignment implicitly, via deformable convolutions or related sampling operators~\cite{dudhane2022burst,mehta2022adaptive,luo2022bsrt,mehta2023gated,dudhane2023burstormer}. Our GAS approach follows the same high-level principle preserving native-view content, but applies it differently. Specifically, we keep each frame in its native viewpoint for local convolutional extraction, and use correspondence only to define the sequence serialization paths used by Mamba state updates. Alignment therefore serves purely as an indexing mechanism for burst message passing, rather than as a preprocessing step that alters features before local extraction. This distinguishes GAS from pre-warping approaches, from methods such as KPN~\cite{mildenhall2018burst} that couple alignment and restoration through learned per-pixel kernels, from approaches that process unaligned frames directly or use only coarse patch-level alignment~\cite{shi2022rethinking}.

\noindentparagraph{Frequency-aware fusion and wavelets} Sub-pixel cues concentrate in high-frequency regions, so multi-frame restoration has long incorporated frequency-aware aggregation, ranging from Fourier-domain accumulation of dominant components~\cite{delbracio2015burst} to wavelet decompositions that fuse high-frequency subbands for sharper results~\cite{huang2017wavelet}. In contrast to using wavelets as an explicit fusion domain, we use wavelet responses to \emph{parameterize} the burst state update, i.e. condition Mamba's selective parameters on wavelet-derived features, biasing aggregation toward HF evidence without deepening the burst stream.

\noindentparagraph{Mamba for BISR} remains relatively underexplored. QMambaBSR~\cite{di2024qmambabsrburstimagesuperresolution} incorporates Mamba-style state updates within a query-based fusion framework, where the \emph{keyframe acts as the sole query} to aggregate information from the burst. BurstMamba is architecturally and operationally distinct, sharing only the use of Mamba state updates. We highlight three central differences: {(1) Compute allocation and modularity:} QMambaBSR performs heavy burst processing within its fusion blocks, whereas BurstMamba explicitly separates a high-capacity \emph{keyframe SR backbone} from a \emph{shallow burst prior stream}. This enables a large reconstruction backbone while keeping burst processing as a small per-frame incremental cost, yielding favorable scaling with burst length. {(2) Transformer dependence:} QMambaBSR still relies on transformer components within its fusion machinery, while BurstMamba is transformer-free and uses SSM routing in both the single-image and burst streams. {(3) How burst evidence is routed into the keyframe:} QMambaBSR aggregates burst evidence through keyframe-query-driven fusion, whereas BurstMamba uses burst sequence SSM routing to distill \emph{per-pixel keyframe residual priors} from the burst and injects them stage-wise into the keyframe stream as residual additions.

\noindentparagraph{Asymmetric processing} Asymmetric compute allocation has been explored in other fields such as reference-based super-resolution (RbSR)~\cite{cao2022reference}, where a high-resolution or telephoto reference provides explicit texture to transfer into the target. In that setting, asymmetry is natural: the reference already contains the desired detail. BISR presents a harder version of this problem: all frames are low-resolution, and useful information lies only in sub-pixel intensity differences across frames, not in any frame's absolute resolution. While prior BISR work has addressed this by applying heavy feature extractors to all frames~\cite{wei2023towards,di2024qmambabsrburstimagesuperresolution}, our central finding is that asymmetric compute allocation remains effective even here, precisely because the burst's role is not to reconstruct but to inform.

\section{Burst Image Super-Resolution with Mamba}
\label{sec:method}

Given a low-resolution burst $\{I^l\}_{l=0}^{L-1}$ with keyframe $I^0$ burst image super-resolution (BISR) reconstructs a high-resolution image $\hat{I}^0_{\mathrm{HR}}$ by aggregating complementary sub-pixel evidence from the remaining frames. Our design is guided by a compute-allocation principle: \emph{the network should dedicate most of its capacity to single-image super-resolution (SISR) of the keyframe, while burst processing should be lightweight and focused on extracting and routing sub-pixel priors.}

We further motivate this design through an information-asymmetry perspective on BISR, discussed in Appendix~\ref{app:sec:info_asymmetry}.

\begin{figure}
  \centering
  \includegraphics[width=\linewidth]{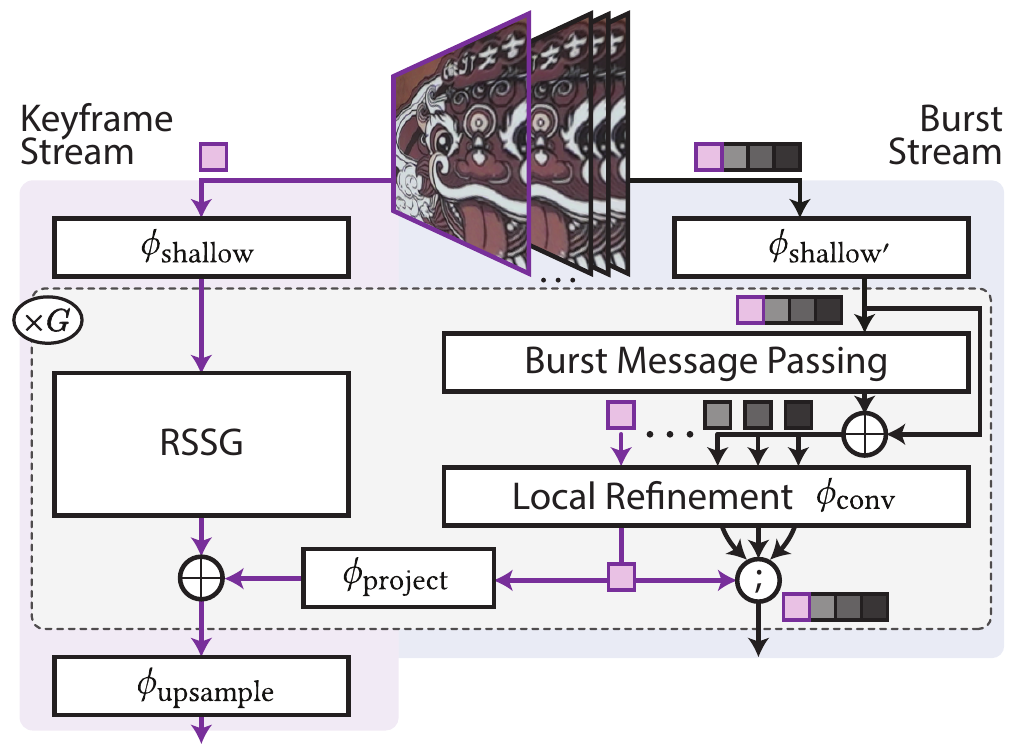}
  \caption{Illustration of BurstMamba's overall pipeline. BurstMamba takes a burst sequence as input and super-resolves the keyframe. The model consists of two key streams: a {\color{keyframepink}keyframe stream} to process only the keyframe for SISR, and a {\color{burstblue}burst stream} to feed sub-pixel priors from the burst into the keyframe stream.}
  \label{fig:pipeline}
\end{figure}

BurstMamba realizes this through two parallel streams (Fig.~\ref{fig:pipeline}): a \textbf{keyframe stream} performing SISR on $I^0$ via a high-capacity backbone, and a lightweight \textbf{burst stream} operating on the entire burst  $\{I^l\}_{l=0}^{L-1}$ to produce a \textit{burst prior} to aid the keyframe reconstruction. At each of the $G$ stage boundaries, the burst prior is added to the keyframe stream, with the coupling being strictly one-way. After $G$ stages, a reconstruction head $\phi_{\mathrm{upsample}}$ produces the final $\hat{I}^0_{\mathrm{HR}}$. The full forward pass is summarized in Algorithm~\ref{alg:burstmamba}.

\begin{algorithm}[t]
  \caption{BurstMamba forward pass}
  \label{alg:burstmamba}
  \begin{algorithmic}[1]
    \REQUIRE Burst $\{I^l\}_{l=0}^{L-1}$, keyframe $I^0$
    \STATE $\mathbf{s}_0 \leftarrow \phi_{\mathrm{shallow}}(I^0)$
    \STATE $\mathbf{f}^l_0 \leftarrow \phi_{\mathrm{shallow'}}(I^l)$ for all $l = 0, \dots, L-1$
    \FOR{$g = 0, \dots, G-1$}
      \STATE $\{\mathbf{f}^l_{g+1}\} \leftarrow \mathrm{BurstStream}_g(\{\mathbf{f}^l_g\})$
      \STATE $\hat{\mathbf{s}}_{g+1} \leftarrow \mathrm{KeyframeStream}_g(\mathbf{s}_g)$
      \STATE $\mathbf{s}_{g+1} \leftarrow \hat{\mathbf{s}}_{g+1} + \phi_{\mathrm{project}}(\mathbf{f}^0_{g+1})$
    \ENDFOR
    \STATE $\hat{I}^0_{\mathrm{HR}} \leftarrow \phi_{\mathrm{upsample}}(\mathbf{s}_G)$
  \end{algorithmic}
\end{algorithm}

\subsection{Keyframe Stream: Single-Image Super-Resolution}

The keyframe stream starts by projecting the keyframe image onto a $C$ dimensional representation via a shallow feature extractor \mbox{$\mathbf{s}_0=\phi_{\mathrm{shallow}}(I^0)$}, followed by a stack of Residual State-Space Groups (RSSG)~\cite{guo2024mambair}. At each stage, RSSG flattens the keyframe features along four spatial serialization patterns to construct 1D token sequences, then models dependencies between tokens via selective SSMs to progressively refine the keyframe representation:
\begin{equation}
  \hat{\mathbf{s}}_{g+1} = \mathrm{RSSG}_g(\mathbf{s}_g).
\end{equation}
This constitutes the main computational backbone of BurstMamba, and is where the bulk of the network capacity is concentrated. We refer to MambaIR~\cite{guo2024mambair} for a full description of RSSG. We illustrate the keyframe stream in Fig.~\ref{fig:pipeline}-left highlighted in pink.

\subsection{Burst Stream: Burst Prior Extraction}
\label{sec:temporal_s6}

The burst stream processes the full burst at each stage to distill cross-frame evidence into an updated keyframe prior. It operates at LR resolution throughout, keeping the computational cost independent of the SR scale factor. The burst stream begins by mapping each burst frame to a reduced-dimensional representation with $C' < C$ via a shallow feature extractor, \mbox{$\mathbf{f}^l=\phi_{\mathrm{shallow'}}(I^l),\ \forall l$}. As illustrated in Fig.~\ref{fig:pipeline} (right, highlighted in blue), processing proceeds in two stages: \textit{burst message passing}, which performs inter-frame communication; and \textit{local refinement}, which extracts intra-frame features.

\paragraph{Burst message passing.} For each spatial location $\mathbf{u} \in \mathbb{Z}^2$, the features across all $L$ frames are serialized into a token sequence:
\begin{equation}
  \label{eq:serialize}
  \mathbf{x}_g= \left[ \mathbf{f}^l_g[\mathbf{u}] \right]_{l=0}^{L-1}
\end{equation}
which is then processed by a bidirectional S6 operator to aggregate cross-frame information:
\begin{equation}
  \label{eq:aggregate}
  \tilde{\mathbf{x}}_{g} =
  \mathrm{S6}_{\mathrm{fwd}}(\mathbf{x}_{g}) +
  \mathrm{S6}_{\mathrm{rev}}(\mathbf{x}_{g}).
\end{equation}
We provide preliminaries of Mamba and the S6 operator in Appendix~\ref{app:sec:preliminaries}. The bidirectional S6 operator preserves sequence length, producing one output token per frame \newtext{per location.} Stacking $\tilde{\mathbf{x}}_{g}$ over all spatial locations $\mathbf{u}$ yields a feature volume $\mathbf{r}^l_g \in \mathbb{R}^{C' \times H \times W}$ per frame $l$, which is treated as a residual on the initial features.

\begin{figure}[t]
  \centering
  \includegraphics[width=\columnwidth,trim=0 0 80.58bp 0,clip]{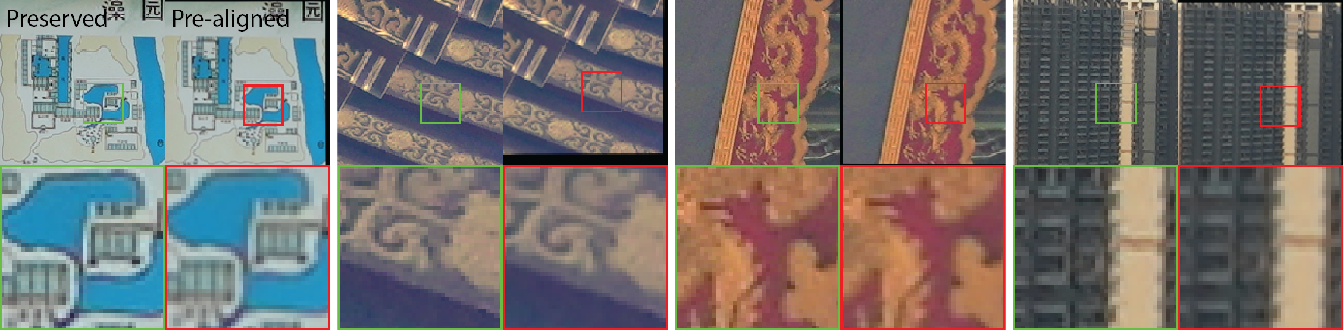}
  \caption{\label{fig:dbs_visual1} Comparison between a burst frame in \textit{native} view (left) and its version aligned to the keyframe (right). Zooming highlights the attenuation of high-frequency cues caused by warping. }
\end{figure}

\begin{figure*}
  \centering
  \includegraphics[width=\linewidth]{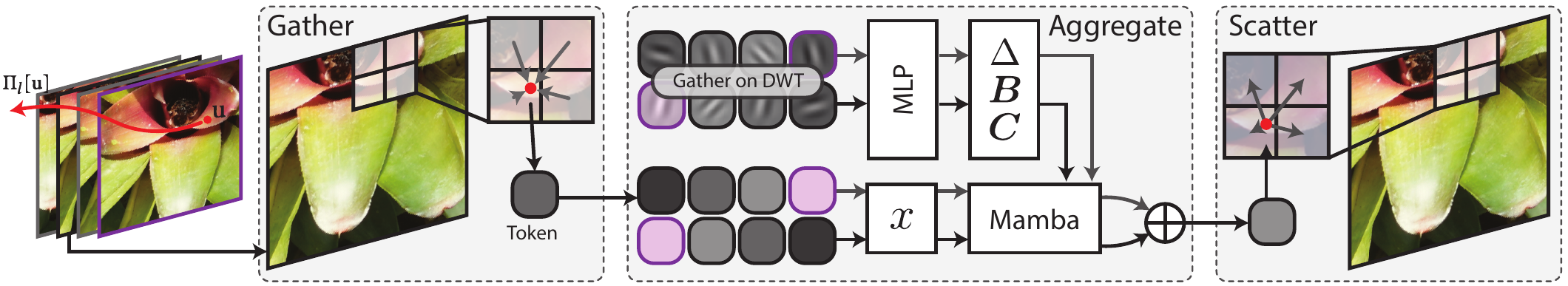}
  \caption{\textbf{Gather \(\rightarrow\) Aggregate \(\rightarrow\) Scatter (GAS) with wavelet-conditioned selective routing.} For each keyframe pixel \(u\), features from all burst frames are \emph{gathered} via bilinear sampling at sub-pixel locations \(\Pi_l[u]\), forming a burst token sequence aligned to the same scene point. This sequence is \emph{aggregated} in keyframe coordinates by a bidirectional Mamba/S6 block, while wavelet features from the same gathered locations parameterize the selective update through \((\Delta,B,C)\) to emphasize high-frequency evidence. Finally, enriched features are \emph{scattered} back to native frame coordinates via inverse correspondences. This allows local refinement to operate on largely unwarped features via a residual connection (omitted for clarity), while leveraging integrated cross-frame priors.}
  \label{fig:gas}
\end{figure*}

\paragraph{Local intra-frame refinement.} Given a residual burst signal $\mathbf{r}^l_g$, each frame is updated by a lightweight convolutional block as:
\begin{equation}
  \label{eq:residual_refine}
  \mathbf{f}^l_{g+1}
  =
  \phi_{\mathrm{conv}}\!\bigl(\mathbf{f}^l_g + \mathbf{r}^l_g\bigr).
\end{equation}
While all refined frames are maintained internally for progressively improved representations in subsequent stages, \newtext{i.e. message passing and residual convolutional refinement are applied so later stages use native-detail and burst priors}, only the updated keyframe feature $\mathbf{f}^0_{g+1}$ is exported as the burst prior and fused into the keyframe stream following a projection layer $\phi_{\mathrm{project}}$ that maps the burst stream's low channel size $C'$ to the keyframe stream's $C$.
\newtext{Analogously to transformers, ``message passing" acts as self-attention-like inter-frame information flow, and ``local refinement" as a feed-forward-like intra-frame feature extraction process, jointly achieving complementary halves of one mechanism.}

While straightforward, this baseline formulation raises a practical challenge: because frames are not aligned, the token sequence $\mathbf{x}_{g}$ in Eq.~\ref{eq:serialize} typically mixes features from different physical scene points across the burst, so the aggregation step in Eq.~\ref{eq:aggregate} is not guaranteed to operate on clean cross-frame correspondences. This can lead to two regimes: either the convolutional refinement layers compensate implicitly, reorganizing or diffusing local evidence so later message passing stages receive more useful signals, or the selective message passing operator learns to suppress misaligned content, making part of the burst interaction redundant Pre-aligning frames alleviates this ambiguity, but introduces warping, which smooths high-frequency content as illustrated in Fig.~\ref{fig:dbs_visual1} and leaves the local refinement step in Eq.~\ref{eq:residual_refine} operating on degraded frame content.

\subsection{Gather $\rightarrow$ Aggregate $\rightarrow$ Scatter (GAS)}
\label{sec:dbs}

We address this challenge through the \emph{Gather $\rightarrow$ Aggregate $\rightarrow$ Scatter} procedure shown in Fig.~\ref{fig:gas}, which extends the naive \textit{burst message passing} from Sec.~\ref{sec:temporal_s6}. Rather than using correspondences for direct warping, we use them \textit{only} to facilitate cross-frame message passing.

\paragraph{Correspondence maps.} For each non-keyframe $l$, we estimate a dense displacement map $\delta^{0 \rightarrow l} \in \mathbb{R}^{2 \times H \times W}$, where $\delta^{0 \rightarrow l}[\mathbf{u}]$ gives the sub-pixel displacement from keyframe coordinate $\mathbf{u} \in \mathbb{Z}^2$ to its corresponding location in frame $l$, mapping it to a sub-pixel location $\Pi_l[\mathbf{u}] \in \mathbb{R}^2$ in frame $l$:
\begin{equation}
  \Pi_l[\mathbf{u}] = \mathbf{u} + \delta^{0 \rightarrow l}[\mathbf{u}],
\end{equation}
with $\Pi_0[\mathbf{u}] = \mathbf{u}$ for the keyframe itself. The inverse map $\Pi_l^{-1}$ is obtained analogously from $\delta^{l \rightarrow 0}$.

\paragraph{Gather.} Instead of directly indexing $\mathbf{f}^l_g[\mathbf{u}]$, we sample each frame at the location corresponding to keyframe pixel $\mathbf{u}$:
\begin{equation}
  \label{eq:gather}
  \mathbf{x}_g= \textrm{Gather}(\mathbf{f}^l_g,\mathbf{u})=\left[ \mathcal{W}(\mathbf{f}^l_g,\,
  \Pi_l[\mathbf{u}]) \right]_{l=0}^{L-1},
\end{equation}
where $\mathcal{W}$ denotes bilinear sampling. This yields a per-pixel token sequence $\mathbf{x}_g$ of length $L$ where each token corresponds to the same physical scene point across frames.

\paragraph{Aggregate.} The gathered token sequence $\mathbf{x}_{g}$ is processed by the bidirectional S6 operator (Eq.~\ref{eq:aggregate}), yielding an enriched sequence $\tilde{\mathbf{x}}_{g}$ expressed in keyframe coordinates. Stacking over all spatial locations yields the feature volume ${\mathbf{r}}^l_g \in \mathbb{R}^{C' \times H \times W}$ per frame $l$.

\paragraph{Scatter.} After aggregation, the enriched signal is expressed in keyframe coordinates. We do not replace the native burst frame features with this signal; instead, we scatter it back to each frame's native coordinate system and add it \emph{residually}. This is important: the scattered component carries the cross-frame information obtained through correspondence-aware aggregation, while the residual pathway preserves the original native-view features that retain the local sub-pixel detail. Thus, any smoothing introduced by bilinear sampling affects only the transferred component, not the underlying frame representation on which the subsequent local refinement operates.

Formally, we back-project each volume via the inverse map $\Pi_l^{-1}[\mathbf{v}]$, where $\mathbf{v} \in \mathbb{Z}^2$ denotes a coordinate in frame $l$'s space:
\begin{equation}
  \tilde{\mathbf{f}}^l_g[\mathbf{v}] = {\mathbf{f}}_g^l +
  \mathcal{W}\!\left({\mathbf{r}}^l_g,\, \Pi_l^{-1}[\mathbf{v}]\right),
  \quad l = 0, \dots, L-1.
\end{equation}
The local intra-frame refinement (Eq.~\ref{eq:residual_refine}) then proceeds as before.

\newtext{Intuitively, \textit{Gather} is a temporary lookup that brings matching points together for aggregation. \textit{Scatter} puts the signal back into each frame's native view; otherwise, non-keyframe features remain warped into the keyframe view, effectively pre-aligned. Thus, GAS does not permanently align frames: it borrows aligned residual information and preserves native details. We provide a visual example of GAS applied to a real image in Appendix~\ref{app:sec:gas}.}

\paragraph{Flow estimation.} GAS is agnostic to how the correspondence field is obtained. When not provided by the dataset, we estimate $\delta$ using a frozen RAFT estimator. We evaluate a homography-based variant in Sec.~\ref{sec:ablation} and find comparable accuracy. Occlusions are not handled explicitly; instead we rely on the S6 operator's selectivity to downweight inconsistent observations during aggregation.

\subsection{Wavelet-conditioned state update ($\psi$S6)}
\label{sec:psi_s6}

Sub-pixel evidence is concentrated in high-frequency regions; low-frequency regions provide less informative priors. Instead of deepening the burst stream (which would redundantly process all $L$ frames), we bias burst selectivity toward high-frequency content by conditioning S6 parameters on wavelet-derived features.

\paragraph{Wavelet features.} For each burst feature we compute a one-level DTCWT and apply \textit{Gather} (Eq.~\ref{eq:gather}). Let $\psi^l_{g}[\mathbf{u}]$ denote the resulting high-frequency feature maps (subbands) associated with frame $l$ at stage $g$, defined for each spatial location $\mathbf{u} \in \mathbb{Z}^2$:
\begin{equation}
  \label{eq:wavelets}
  \psi^l_{g}[\mathbf{u}] = \mathrm{Gather}(\mathrm{DTCWT}(\mathbf{f}^l_g),\mathbf{u}).
\end{equation}
\paragraph{Wavelet-conditioned selectivity.} In standard S6, parameters are conditioned on the input token $x$; here, we instead condition them on wavelet features $\psi$, decoupling content representation from routing decisions. Recall from Sec.~\ref{sec:temporal_s6} that the burst is serialized into a token sequence $\mathbf{x}_g$ (Eq.~\ref{eq:gather}) of length $L$. For a given spatial location $\mathbf{u}$, writing $x_l$ for the $l$-th token of $\mathbf{x}_g[\mathbf{u}]$ and $\psi_l$ for the $l$-th token of $\psi^l_{g}[\mathbf{u}]$, the $\psi$S6 formulation is:
\begin{subequations}
  \label{eq:psis6}
  \begin{align}
    h_l &= \exp(\bm{\Delta}(\psi_{l})\bm{A})\,h_{l-1}+\overline{\bm{B}}(\psi_{l})\,x_l,\\
    y_l &= \bm{C}(\psi_{l})\,h_l,
  \end{align}
\end{subequations}
where $(\bm{\Delta}, \overline{\bm{B}}, \bm{C})$ are predicted by a single linear layer applied to a compact projection of $\psi_l$. This steers burst routing capacity toward high-frequency regions that are most informative for sub-pixel transfer  with minimal added compute.

\begin{table*}[t]
  \centering
  \caption{\label{tab:quantitative_results}Performance comparison of existing burst image SR methods. (Baseline results are taken from the corresponding papers under the official benchmark protocol; '-' indicates the metric is not reported in the original work.)}
  \begin{tabular}{|l|ccc|ccc|ccc|}
    \cline{2-10}
    \multicolumn{1}{c|}{} & \multicolumn{3}{c|}{SyntheticSR} & \multicolumn{3}{c|}{RealBSR-RAW} & \multicolumn{3}{c|}{RealBSR-RGB} \\
    \cline{1-1}
    Method & PSNR$\uparrow$ & SSIM$\uparrow$ & LPIPS$\downarrow$ & PSNR$\uparrow$ & SSIM$\uparrow$ & LPIPS$\downarrow$ & PSNR$\uparrow$ & SSIM$\uparrow$ & LPIPS$\downarrow$ \\
    \hline
    DBSR~\cite{bhat2021deep} & 39.17 & 0.946 & 0.081 & 20.91 & 0.635 & 0.134 & 30.72 & 0.899 & 0.101 \\
    MFIR~\cite{bhat2021deep_mfir} & 41.55 & 0.964 & 0.045 & 21.56 & 0.638 & 0.131 & 30.90 & 0.899 & 0.098 \\
    BIPNet~\cite{dudhane2022burst} & 41.93 & 0.967 & 0.035 & 22.90 & 0.641 & 0.144 & 30.66 & 0.892 & 0.111 \\
    BSRT-L~\cite{luo2022bsrt} & 43.62 & 0.975 & \textbf{0.025} & 22.58 & 0.622 & 0.103 & 30.78 & 0.900 & 0.101 \\
    FBANet~\cite{wei2023towards} & 42.23 & 0.970 & - & 23.42 & 0.677 & 0.125 & 31.01 & 0.898 & 0.102 \\
    SBFBurst~\cite{cotrim2024sbfburst} & 42.19 & 0.968 & 0.036 & - & - & - & 31.07 & 0.903 & 0.096 \\
    Burstormer~\cite{dudhane2023burstormer} & 42.83 & 0.973 & - & 27.29 & 0.816 & - & 31.20 & 0.907 & - \\
    QMambaBSR~\cite{di2024qmambabsrburstimagesuperresolution} & 43.12 & 0.97- & - & 27.56 & 0.820 & - & 31.40 & 0.908 & - \\
    \hline
    BurstMamba (Ours) \ & \textbf{44.51} & \textbf{0.978} & 0.037 & \textbf{28.03} & \textbf{0.832} & \textbf{0.064} & \textbf{33.29} & \textbf{0.929} & \textbf{0.045} \\
    \hline
  \end{tabular}
\end{table*}

\begin{table}[t]
  \centering
  \caption{\label{tab:ablation_components}Ablation study on proposed components starting from the single image super-resolution baseline.}
  \small
  \begin{tabular}{|l|cc|cc|}
    \hline
    Input & GAS & $\psi$S6 & PSNR$\uparrow$ & SSIM$\uparrow$ \\
    \hline
    Single & & & 31.668 & 0.900 \\
    Burst & & &  32.881 & 0.924 \\
    Burst & \ding{51} & & 33.080 & 0.928 \\
    Burst & \ding{51} & \ding{51} & \textbf{33.287} & \textbf{0.929} \\
    \hline
  \end{tabular}
\end{table}

\begin{table}[t]
  \centering
  \caption{\label{tab:ablation_dbs} Ablation study comparing (i) integer and bilinear GAS and (ii) the impact of when to align the input images.}
  \small
  \begin{tabular}{|c|l|l|cc|}
    \cline{2-5}
    \multicolumn{1}{l|}{} & Alignment & Interpolation & PSNR$\uparrow$ & SSIM$\uparrow$ \\
    \hline
    I. & None & - & 32.881 & 0.924 \\
    II. & Pre-Align & Bilinear & 32.965 & 0.926 \\
    III. & Optical-flow GAS& Integer & 32.434 & 0.917 \\
    IV. & Optical-flow GAS& Bilinear & \textbf{33.080} & \textbf{0.928} \\
    V. & Homography GAS& Bilinear & 33.000 & 0.927 \\
    \hline
  \end{tabular}
\end{table}

\section{Experiments}

We evaluate BurstMamba on three public burst image super-resolution benchmarks for $\times4$ enhancement: SyntheticSR~\cite{bhat2021deep} (synthetic RAW$\rightarrow$RGB), RealBSR-RAW~\cite{wei2023towards} (real RAW$\rightarrow$RGB), and RealBSR-RGB~\cite{wei2023towards} (real RGB$\rightarrow$RGB). Unless stated otherwise, we conduct ablations on RealBSR-RGB. For all datasets, we follow the exact same training/evaluation protocol and metric computation used by the corresponding benchmark papers and prior methods. Explicit implementation details are provided in Appendix~\ref{app:sec:impl}, including architectural and training details and discussion of the RAW input representation.

\subsection{Results}
\label{sec:results}

\begin{figure*}[p]
  \centering
  \includegraphics[width=.917\linewidth]{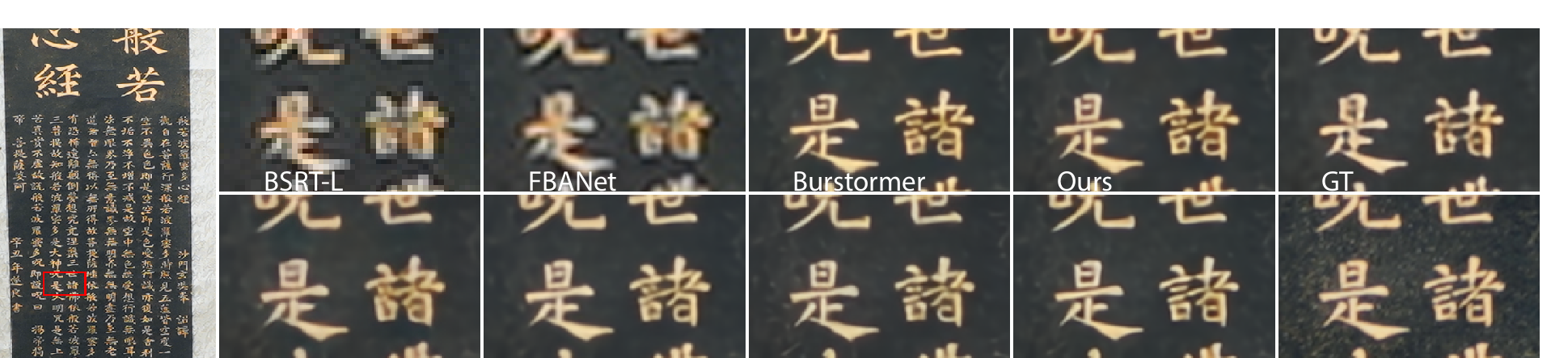}
  \includegraphics[width=.9\linewidth]{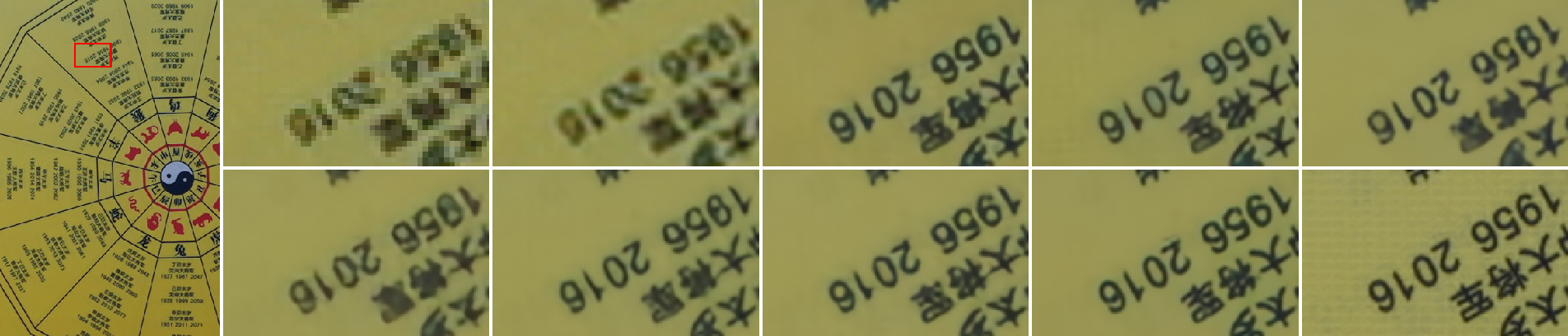}
  \caption{\label{fig:qualitative_result1}Qualitative comparison of different methods and our proposed BurstMamba on the RealBSR-RGB dataset for $\times4$ burst image super-resolution (BISR) with $L=14$.}

  \includegraphics[width=.9\linewidth]{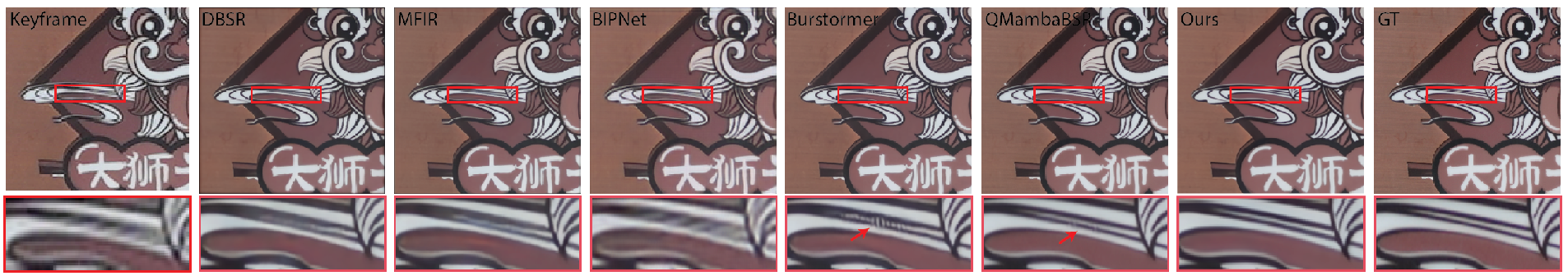}
  \caption{QMambaBSR~\cite{di2024qmambabsrburstimagesuperresolution} does not release code or full-resolution outputs. Nevertheless, its published qualitative exemplars are still informative, as they reflect the representative results the authors chose to showcase. Accordingly, we follow an externally defined evaluation setup: rather than selecting scenes or zoom regions ourselves, we reuse the exemplar image and the exact crop shown in QMambaBSR's figure. This avoids subjective region selection on our side and focuses the comparison on the specific fine structures emphasized in prior work. Under this fixed-exemplar protocol, BurstMamba better preserves the highlighted fine details by maintaining a clean, continuous separation between adjacent edges. For ease of visual inspection, we mark residual artifacts in the Burstormer and QMambaBSR reconstructions with red arrows.}
  \label{fig:qualitative_result_qmbsr}

  \includegraphics[width=.917\linewidth]{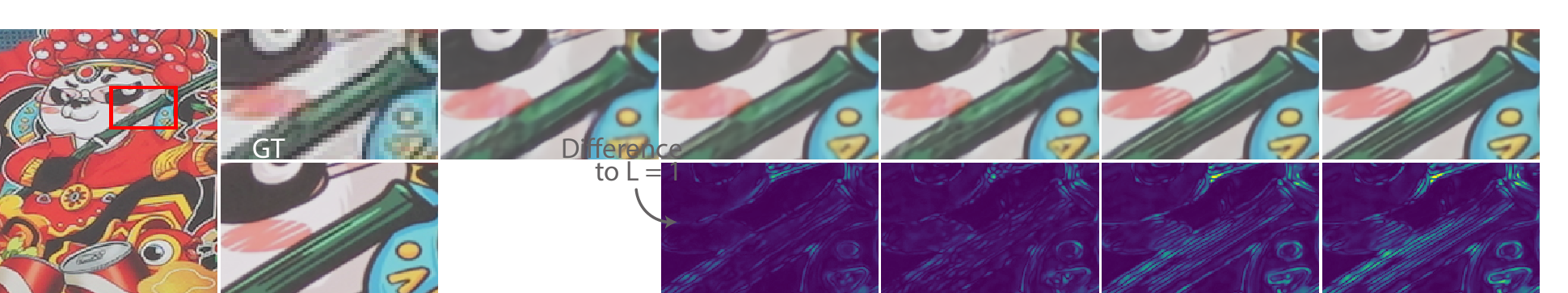}
  \includegraphics[width=.9\linewidth]{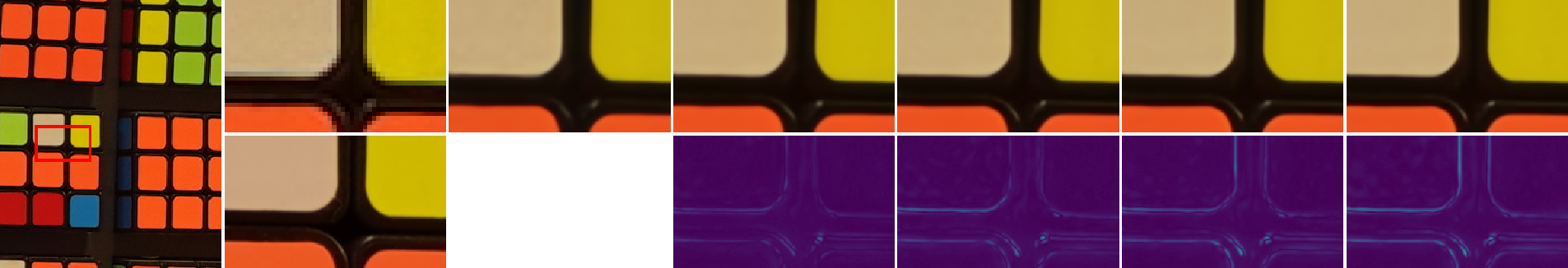}
  \caption{Qualitative results from varying the input burst sequence length (L) for BurstMamba on the RealBSR-RGB dataset. On the top row we illustrate the benefit of increasing the burst length when facing a scene dominated by high frequency details. On the bottom row, we show that single image super-resolution can provide a sufficiently good result when processing a scene with simple structures. The isolated contributions of the burst module show that it only provides high frequency edge information.}
  \label{fig:qualitative_ablation2}
\end{figure*}

We report quantitative results on all three benchmarks in Tab.~\ref{tab:quantitative_results}. BurstMamba achieves consistent improvements over prior methods, with the largest gain on RealBSR-RGB where it improves PSNR by up to 1.886 dB. Fig.~\ref{fig:teaser} and ~\ref{fig:qualitative_result1} show qualitative comparisons on RealBSR-RGB with methods that provide pretrained models or directly their test set results. As seen, BurstMamba better preserves high-frequency structures, such as thin lines and small gaps, by effectively injecting burst-derived sub-pixel priors into the keyframe reconstruction following our compute-separation principle. Further qualitative samples are provided in Appendix~\ref{app:sec:additional-results}.

We further provide a qualitative sample that includes QMambaBSR in Fig.~\ref{fig:qualitative_result_qmbsr}. Here, BurstMamba reconstructs the image more faithfully, specifically yielding a clearer separation between the lines in the zoomed region. As QMambaBSR code is currently unavailable, we include this panel as a reference comparison based on their published results; see the figure caption for further details.

We also report out-of-distribution (OOD) qualitative results on HDR+~\cite{hasinoff2016burst} in Fig.~\ref{fig:ood_qualitative_result1}. Two observations are important. First, HDR+ is substantially noisier than the burst super-resolution benchmarks commonly used in the literature, and BurstMamba consequently exhibits a degree of implicit denoising. Second, this behavior can also be undesirable: when noise patterns partially coincide across frames, their aggregation can reinforce structured artifacts, producing spurious texture in some regions. We clearly demonstrate this limitation of BurstMamba with additional qualitative results in Appendix~\ref{app:sec:additional-results}. These failure cases motivate future work that explicitly models noise in the burst prior pathway, for example, by incorporating a noise-aware degradation model during training and/or adding dedicated denoising components or uncertainty-aware gating to prevent noisy evidence from being over-amplified.

\newtext{We further collect and release 21 qualitative testing samples with $L=14$. Results and download links are provided in Appendix~\ref{app:sec:additional-results}.}

\subsection{Ablation Studies and Discussion}
\label{sec:ablation}

\paragraph{Effect of individual components.} Tab.~\ref{tab:ablation_components} isolates the contributions of the burst stream, GAS, and $\psi$S6. Starting from the SISR baseline, enabling the burst stream yields the dominant improvement (+1.213 dB), indicating that the burst provides complementary sub-pixel evidence beyond the keyframe. Replacing linear sequence serialization (Eq.~\ref{eq:serialize}) with GAS further improves accuracy (+0.199 dB), and wavelet-conditioned updates ($\psi$S6) add an additional +0.207 dB by biasing selective routing toward high-frequency regions.

\paragraph{Scaling with burst length and detachability.} \newtext{A key practical benefit of our design is that a single trained model supports variable burst lengths at test time, with no retraining required: performance increases monotonically as $L$ grows from 1 to 14, with diminishing returns for larger $L$ (see Appendix~\ref{app:sec:additional-ablations}). Notably, the model remains functional even when the burst stream is fully detached ($L$=1), performing within 0.379 dB of an SISR model trained in isolation. Fig.~\ref{fig:qualitative_ablation2} shows burst priors mainly refine high-frequency regions, leaving low-frequency content largely unchanged.}

\paragraph{Gain on a mature benchmark: compute separation.} RealBSR-RGB appears bottlenecked by keyframe reconstruction capacity: a high-capacity SISR backbone alone (our keyframe stream trained in isolation) reaches 31.668 dB (Tab.~\ref{tab:ablation_components}) with $L$=1, which is higher than previously reported BISR results on RealBSR-RGB.
\newtext{This is consistent with the structure of the BISR task itself: since the output is a single reconstructed frame, gains in SISR capacity translate directly into keyframe quality, whereas burst frames need only supply complementary sub-pixel evidence rather than independently bear reconstruction cost. This motivates allocating most model capacity to a keyframe SR backbone, with burst evidence injected additively via a lightweight prior stream. To verify that this allocation choice, rather than total capacity alone, is responsible for the resulting gain, we conduct an additional ablation constructing a model with the same total GFLOPs, but with compute distributed approximately 50/50 between the SISR backbone and the burst module (see Appendix~\ref{app:sec:impl} for details). Under this setting, we obtain 30.32~dB PSNR and 0.8936 SSIM on RealBSR, a significant drop that supports our central design principle: concentrating capacity on the keyframe-only branch, while using burst frames primarily for prior extraction, is more effective than distributing compute evenly across branches. }

\paragraph{Where to align and the importance of sub-pixel interpolation.} Tab.~\ref{tab:ablation_dbs} compares different alignment strategies and correspondence interpolation. Bilinear sampling is essential: rounding correspondence maps to integers discretizes sub-pixel motion and introduces collisions/holes in the mapping, which degrades sub-pixel cues and harms performance (integer GAS vs bilinear GAS). Moreover, applying alignment only within serialization (GAS) outperforms pre-warping frames/features before local extraction, supporting our claim that preserving native-view frame content benefits intra-frame convolutions responsible for fine-detail extraction.

\paragraph{RAFT vs. a global motion model.} In GAS, we use a dense correspondence field to index sub-pixel samples during serialization; as stated in the method section, this can be provided by RAFT or by a lightweight global motion model (homography). On RealBSR-RGB, homography-based motion modeling achieves comparable accuracy (33.00 vs 33.08 PSNR see Tab.~\ref{tab:ablation_dbs}.V.) with substantially lower correspondence overhead.

\paragraph{Complexity and scaling.} We report complexity following set precedent~\cite{di2024qmambabsrburstimagesuperresolution} on $48{\times}48$ LR inputs (see Fig.~\ref{fig:teaser}). The SISR backbone has 20.6M parameters and costs 63 GFLOPs. The burst stream adds only 0.79M parameters and incurs an additional 1.3 GFLOPs {per extra burst frame}, yielding total compute of approximately $63 + 1.3(L-1)$ GFLOPs. These reported GFLOPs/parameters account for the BurstMamba network only and exclude correspondence estimation (e.g., RAFT or homography) as standard in BISR literature, which is seen as an external pre-processing step. \newtext{At full image resolution, cost scales linearly with both image area and $L$.}

For completeness, the additional cost depends strongly on the choice of correspondence estimator. Under our settings, a RAFT-based implementation adds approximately 0.7 GFLOPs or 3.6 GFLOPs based on the RAFT size, while a homography-based strategy (SIFT + FLANN + RANSAC) adds only about 0.002--0.02 GFLOPs per additional burst image depending on the number of RANSAC iterations, bringing the total compute for $L=14$ to 89, 126.7, and 79.9 GFLOPs respectively. Despite this gap in cost, we found no measurable difference between different sized RAFT models on any of the three evaluated datasets, likely because motion in these benchmarks is dominated by approximately global homographic changes. The homography variant is therefore effectively negligible relative to the BurstMamba network, yet still achieves accuracy close to RAFT-based GAS (Tab.~\ref{tab:ablation_dbs}). Overall, this makes the homography-based configuration a particularly practical operating point, preserving BurstMamba's favorable system-level scaling while retaining most of the benefit of correspondence-aware burst aggregation.

\begin{figure*}[p]
  \centering
  \includegraphics[width=.88\linewidth]{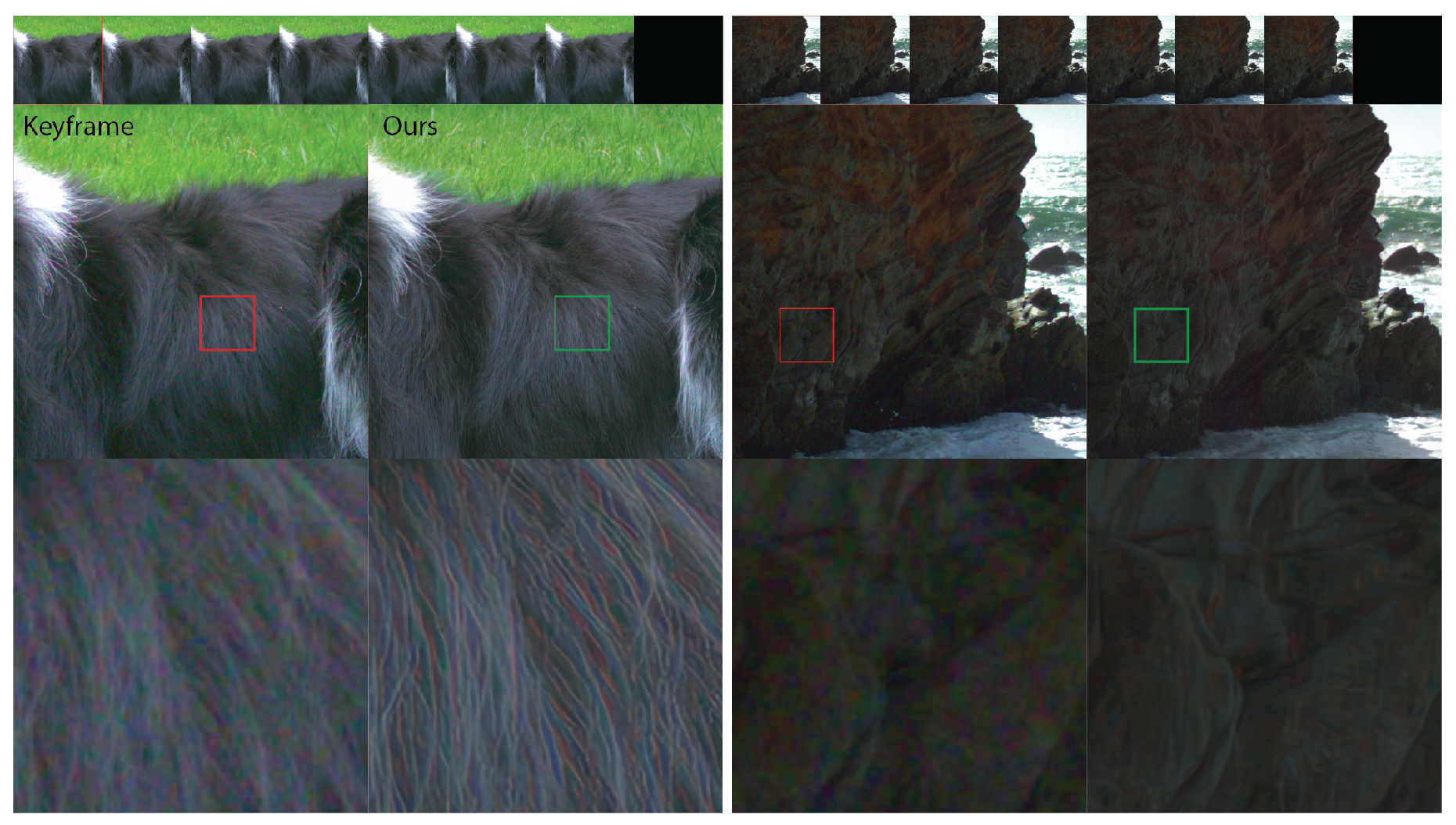}
  \caption{Qualitative out-of-distribution results of BurstMamba trained on RealBSR-RGB dataset with $L=14$ and evaluated on the HDR+ burst photography dataset with $L=7$.}
  \label{fig:ood_qualitative_result1}
  \includegraphics[width=.88\linewidth]{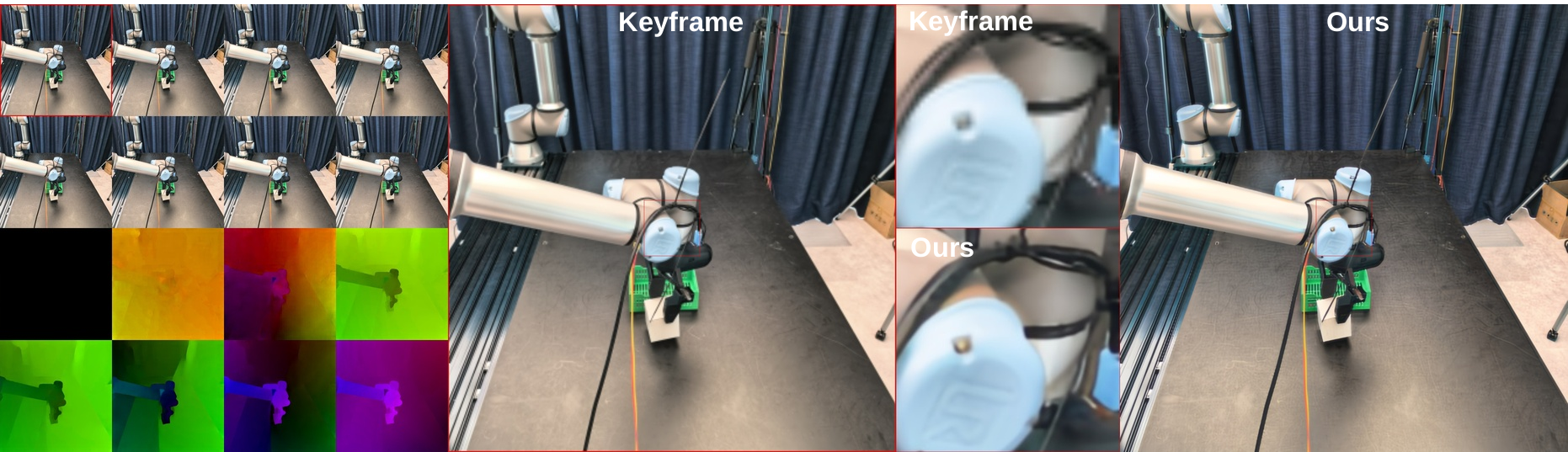}
  \includegraphics[width=.88\linewidth]{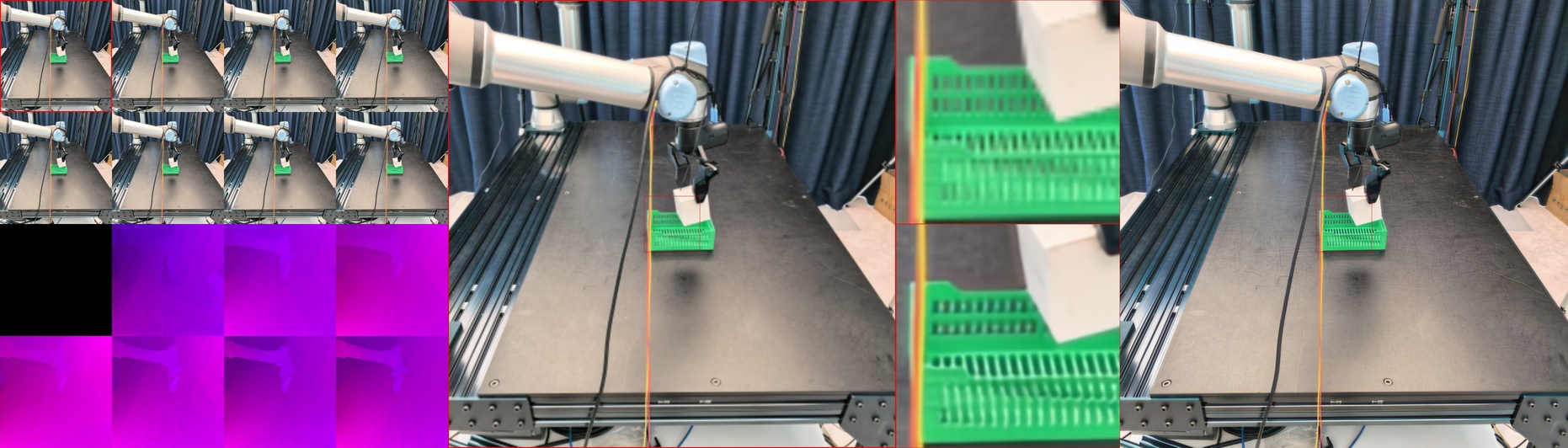}
  \caption{Qualitative OOD results of BurstMamba evaluated on a self-collected dataset. We show that our model can leverage optical flow cues with GAS to reconstruct the fine details (top) on the moving salient object (bottom) behind the moving object.}
  \label{fig:ood_qualitative_result2}
\end{figure*}

\paragraph{Testing under occlusions.} While the results above indicate that simple correspondence models are sufficient on current BISR benchmarks, these datasets are largely dominated by near-global motion and do not expose the model to strong occlusions or pronounced local motion during training. To probe this regime, we additionally evaluate BurstMamba on further OOD sequences captured in front of textured objects with local motion using a robotic arm via a mobile phone, and show representative examples in Fig.~\ref{fig:ood_qualitative_result2}. We show that the method reconstructs fine structures on and behind the moving salient robotic arm, despite not being trained explicitly for local motion. We further discuss limitations under occlusions and optical flow failure in Appendix~\ref{app:sec:additional-ablations}.

\section{Conclusion}

We presented BurstMamba, a burst image super-resolution architecture built on a simple principle: allocate most compute to reconstructing the keyframe, and use the remaining burst frames primarily to extract complementary sub-pixel priors. Accordingly, BurstMamba decouples BISR into a high-capacity Mamba-based keyframe SR stream and a lightweight burst prior stream coupled only through stage-wise residual injection. We further introduced \emph{Gather $\rightarrow$ Aggregate $\rightarrow$ Scatter} (GAS), which uses correspondence only for cross-frame message passing while preserving native-view features through a residual formulation, and the wavelet-conditioned state update (\(\psi\)S6), which biases selective routing toward high-frequency regions without deepening the burst branch.

Across SyntheticSR, RealBSR-RAW, and RealBSR-RGB, BurstMamba achieves strong results and scales favorably with burst length, supporting the core hypothesis that keyframe-centric compute allocation is an effective design principle for BISR. Extensive ablations and analysis further validate the contribution of each component, including the compute-separation design, GAS, the wavelet-conditioned update, burst-length scaling, and the practical trade-off between RAFT- and homography-based correspondence estimation. Our discussion highlights important limitations under correspondence errors, strong occlusions, and noisy burst observations to motivate future work on more robust burst prior modeling.

\bibliographystyle{ACM-Reference-Format}
\bibliography{references}

\clearpage
\appendix
\newcommand{\comparisonheaders}{\begin{tabular*}{\linewidth}{@{\extracolsep{\fill}}ccccc@{}}
  \textbf{Input} & \textbf{Flow} & \textbf{Keyframe} &
  \textbf{FBANet} & \textbf{Ours}
\end{tabular*}\par}

\section{Preliminaries: Selective State-Space Models}
\label{app:sec:preliminaries}

An SSM maps an input sequence $\{x_l\}$ to outputs $\{y_l\}$ via a latent state $h_l$:
\begin{equation}
  \label{app:eq:ssm_disc}
  h_l=\overline{\bm{A}}\,h_{l-1}+\overline{\bm{B}}\,x_l,\qquad
  y_l=\bm{C}\,h_l,
\end{equation}
where $(\overline{\bm A},\overline{\bm B},\bm C)$ are discrete-time parameters. Mamba's selective SSM (S6) introduces content-dependent routing by making the discretization and readout depend on the current input:
\begin{subequations}
  \label{app:eq:s6}
  \begin{align}
    h_l &= \exp(\bm{\Delta}(x_l)\bm{A})\,h_{l-1}+\overline{\bm{B}}(x_l)\,x_l,\\
    y_l &= \bm{C}(x_l)\,h_l.
  \end{align}
\end{subequations}
For a full derivation, please refer to Mamba~\cite{gu2023mamba}.

\section{Discussion: Information-Asymmetry Perspective on Compute Separation}
\label{app:sec:info_asymmetry}

We provide a theoretical perspective on why an asymmetric allocation of computation between the keyframe and burst frames may be effective for burst image super-resolution (BISR). Let $\{I_l\}_{l=0}^{L-1}$ denote a burst of low-resolution observations, with $I_0$ designated as the keyframe, and let $\hat{I}_0^{\mathrm{HR}}$ denote the reconstructed high-resolution output. We denote by $Y$ the unknown high-resolution image aligned with $I_0$.

\paragraph{Output-aligned reference and conditional information.} Although all burst frames are observations of the same underlying scene, the reconstruction target $Y$ is defined in the coordinate system of the keyframe $I_0$. As a result, $I_0$ plays a distinct role: it provides a direct, output-aligned observation of $Y$, whereas the remaining frames $\{I_l\}_{l=1}^{L-1}$ are displaced and degraded measurements whose information must be transferred into the keyframe coordinate system.

From an information-theoretic perspective, the contribution of the burst is naturally expressed as conditional information. The reduction in uncertainty about $Y$ after observing the burst in addition to the keyframe is given by:
\begin{equation}
  H(Y \mid I_0) - H(Y \mid I_0, B)
  =
  I(Y; B \mid I_0),
\end{equation}
where $B = \{I_l\}_{l=1}^{L-1}$. This quantity captures the information about $Y$ that is not already available from the keyframe alone. Importantly, this formulation does not assume that auxiliary frames are less informative in an absolute sense; rather, it distinguishes between information that is already aligned with the output (through $I_0$) and information that must be incorporated conditionally.

\paragraph{A conditional-residual interpretation.}
One way to interpret the estimation problem is through a decomposition into a keyframe-based prediction and a burst-conditioned refinement. Using conditional expectations, we can write:
\begin{equation}
  \mathbb{E}[Y \mid I_0, B]
  =
  \mathbb{E}[Y \mid I_0]
  +
  \Big(\mathbb{E}[Y \mid I_0, B] - \mathbb{E}[Y \mid I_0]\Big),
\end{equation}
where the second term represents the improvement obtained by incorporating the burst. This decomposition is not unique, but provides a convenient lens: the first term captures the estimate obtained from the output-aligned keyframe, while the second term captures the additional information contributed by the burst.

In practice, this decomposition motivates the design used in BurstMamba. The keyframe provides an output-aligned estimate of the target, while the burst contributes additional information that refines this estimate through cross-frame evidence. From this perspective, the role of the burst is not to independently reconstruct the high-resolution image, but to supply information that is useful beyond the keyframe, i.e., information related to the conditional term $I(Y; B \mid I_0)$.

\paragraph{Implications for compute allocation.} This interpretation highlights a potential asymmetry in the BISR task. The mapping $I_0 \mapsto Y$ is a highly ill-posed inverse problem requiring the synthesis of high-frequency image structure, as widely studied in super-resolution \cite{liang2021swinir, guo2024mambair}. In contrast, the burst contributes by reducing the remaining uncertainty in the keyframe estimate through conditional information. Since the final output is defined in the coordinate system of $I_0$, the dominant reconstruction burden lies in modeling this mapping, while the burst pathway provides complementary evidence that refines it.

If the information contributed by the burst can be represented with lower complexity than the full reconstruction mapping, it is reasonable to allocate more capacity to the keyframe pathway than to the burst pathway. We emphasize that this is not a claim of optimality, but a theoretically motivated hypothesis: an asymmetric allocation of computation, where a high-capacity keyframe reconstruction is combined with a lightweight burst-processing pathway, may provide a favorable trade-off between capacity and efficiency. BurstMamba instantiates this hypothesis by concentrating model capacity on the keyframe while using a lightweight burst stream to extract and route complementary information.

\section{GAS}
\label{app:sec:gas}

\begin{figure}
  \centering
  \includegraphics[width=\columnwidth]{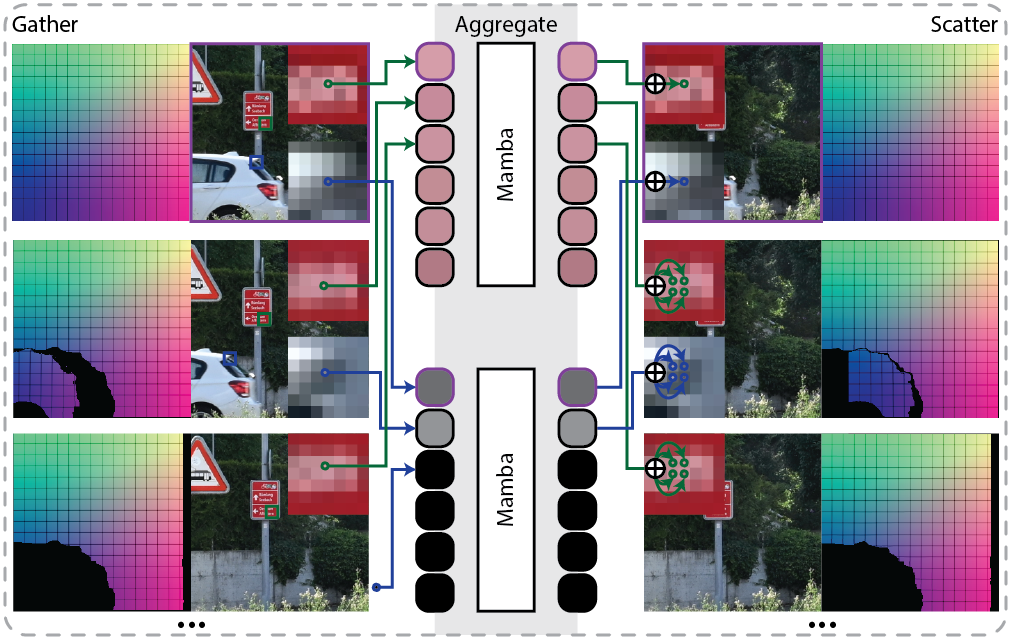}
  \caption{\newtext{GAS temporarily gathers burst-frame features into the keyframe coordinate system for correspondence-aware message passing, then scatters the updated information back to each frame's native coordinates so the burst stream preserves local, unaligned frame structure.}}
  \label{app:fig:sup_gas}
\end{figure}

\newtext{As illustrated in Fig.~\ref{app:fig:sup_gas}, GAS should be understood as temporary alignment for communication, not permanent alignment for reconstruction. For a keyframe location, the gather step follows the estimated correspondence into each burst frame and samples the feature that observes the same scene point. This places the keyframe and burst features into a common coordinate system, so the network can exchange information between genuinely corresponding content rather than between pixels that merely share the same image index. However, if the gathered features were kept in the keyframe grid, the burst stream would effectively become a pre-aligned stack and lose each frame's native local structure, including the sub-pixel offsets and frame-specific sampling patterns that make burst SR useful. GAS therefore scatters the updated message back to the original burst-frame coordinates. In this way, correspondence is used only at the moment of cross-frame interaction, while each burst branch continues to evolve in its own native coordinate system.}

\section{Implementation Details}
\label{app:sec:impl}

\paragraph{Architecture.}
BurstMamba is composed of a high-capacity keyframe stream for and a lightweight burst stream for burst prior extraction (see Fig.~\ref{fig:pipeline}). $\phi_{\mathrm{shallow}}$ projects the keyframe to  $C{=}180$ channels, whereas $\phi_{\mathrm{shallow'}}$ projects the burst to $C'{=}45$. There are a total of $G{=}6$ stages. We use an $\times4$ upsampling head $\phi_{\mathrm{upsample}}$ following MambaIR~\cite{guo2024mambair}. For DTCWT, we use \texttt{pytorch\_wavelets.DTCWTForward} with $J{=}1$, \texttt{biort="near\_sym\_b"}, \texttt{qshift="qshift\_b"}~\cite{pytorch_wavelets_github} followed by an interpolation operation to preserve resolution.

\paragraph{RAW input representation and fairness.} For RAW$\rightarrow$RGB benchmarks, we feed the dataset RAW observation in a \emph{mosaicked RGGB} representation and do not perform explicit demosaicing as a preprocessing step. This choice is made for architectural consistency across RGB$\rightarrow$RGB and RAW$\rightarrow$RGB settings: it reduces the required changes to the model to a single input-channel adjustment, whereas several prior architectures require more substantial changes (e.g., additional upsampling depth) when switching between RGB and RAW input resolutions/packings. Importantly, the mosaicked RAW does not provide extra information relative to standard RAW packings; if anything, it is a more constrained input representation that requires the network to learn demosaicing jointly with super-resolution. All comparisons are made using the same public datasets, the same official splits, and the same evaluation protocol as prior work.

\paragraph{Training schedule.} We train each benchmark from scratch in two stages. First, we train the keyframe SR stream for 150k iterations using $40{\times}40$ LR patches. Second, we enable the burst stream and train the complete model for an additional 250k iterations using $30{\times}30$ LR patches. We use a fixed batch size of 20, burst length $L{=}14$, AdamW, and learning rate $10^{-4}$. We train using an $\ell_1$ reconstruction loss.

\paragraph{Ablation setting for the compute separation baseline.} By tuning the branch-specific channel scaling factors, we reduced the capacity of the keyframe SR backbone to 41.8 GFLOPs and increased the capacity of the burst branch to 2.0 GFLOPs per additional burst frame, yielding a rebalanced variant in which the total GFLOPs remained nearly unchanged while the compute of the two branches became approximately matched.

\begin{figure*}[th!]
  \centering
  \includegraphics[width=1.017\linewidth]{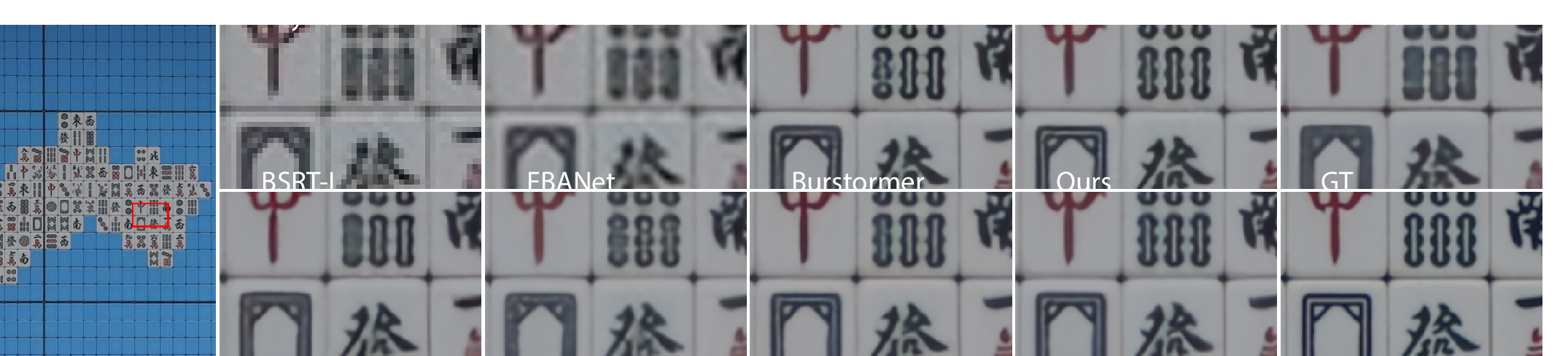}
  \includegraphics[width=\linewidth]{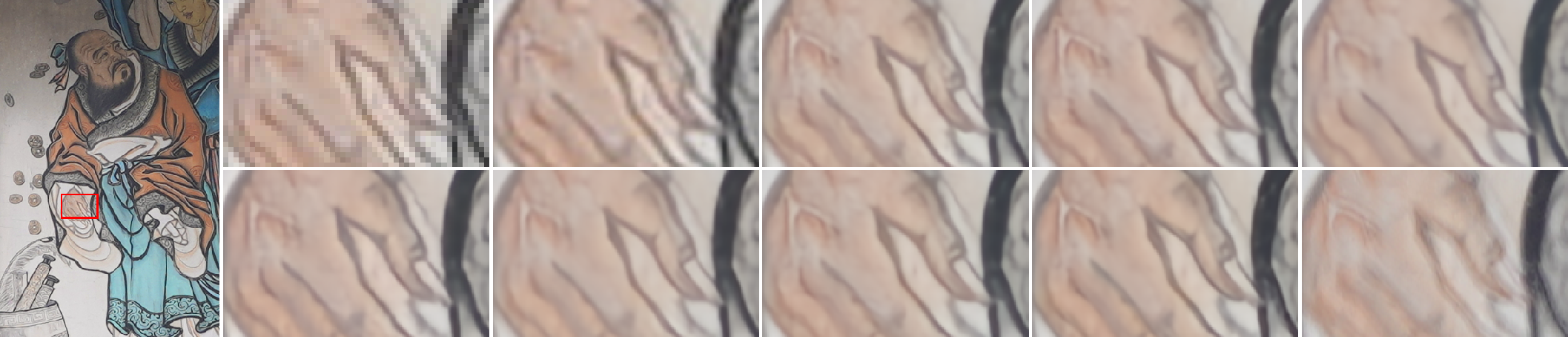}
  \includegraphics[width=\linewidth]{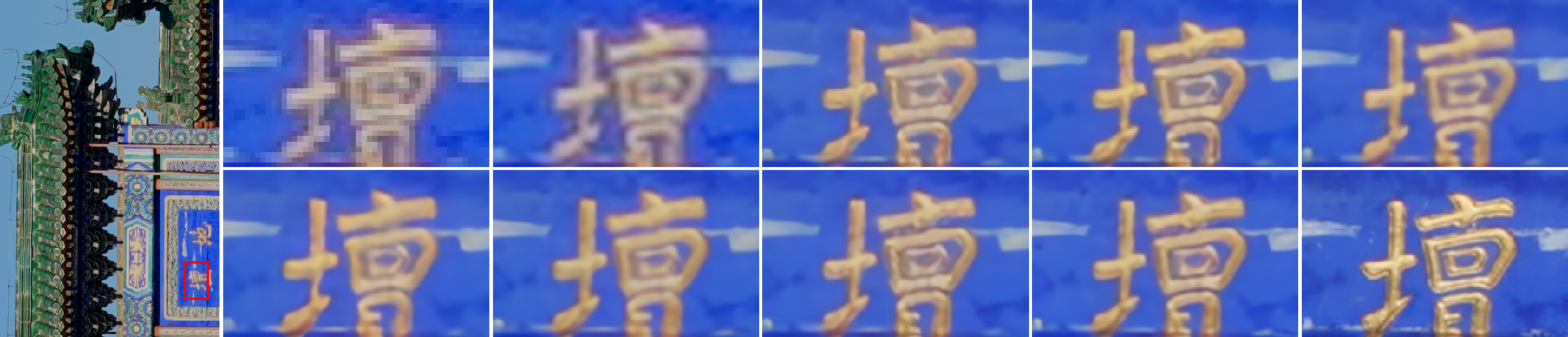}
  \caption{Qualitative comparison of different methods and our proposed BurstMamba on the RealBSR-RGB dataset for $\times4$ burst image super-resolution (BISR) with $L=14$. In row 2, we show that all compared BISR methods present a better reconstruction than the given blurry ground truth image.}
  \label{app:fig:sup_qualitative_result1}
\end{figure*}

\section{Additional Results}
\label{app:sec:additional-results}

\begin{figure*}[t]
  \centering
  \includegraphics[width=1.017\linewidth]{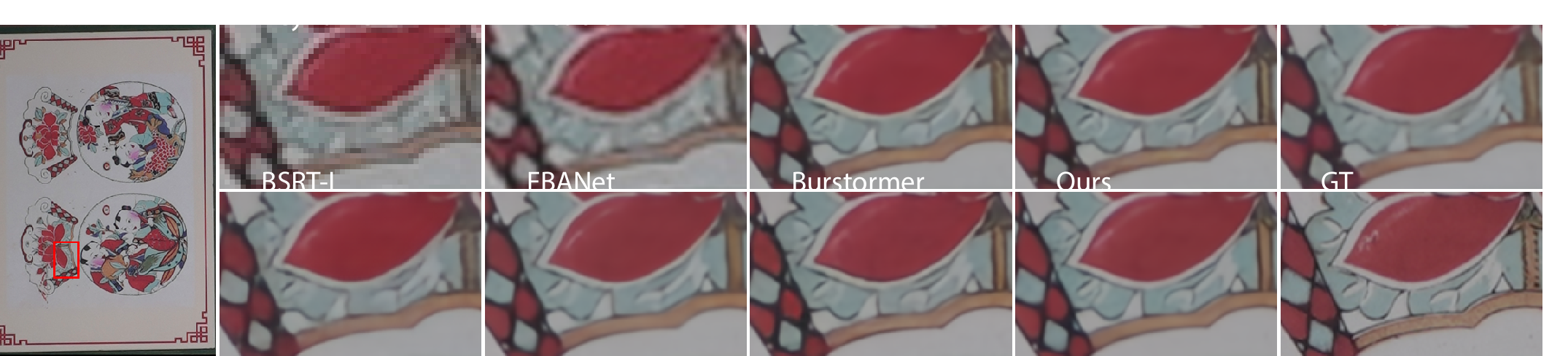}
  \includegraphics[width=\linewidth]{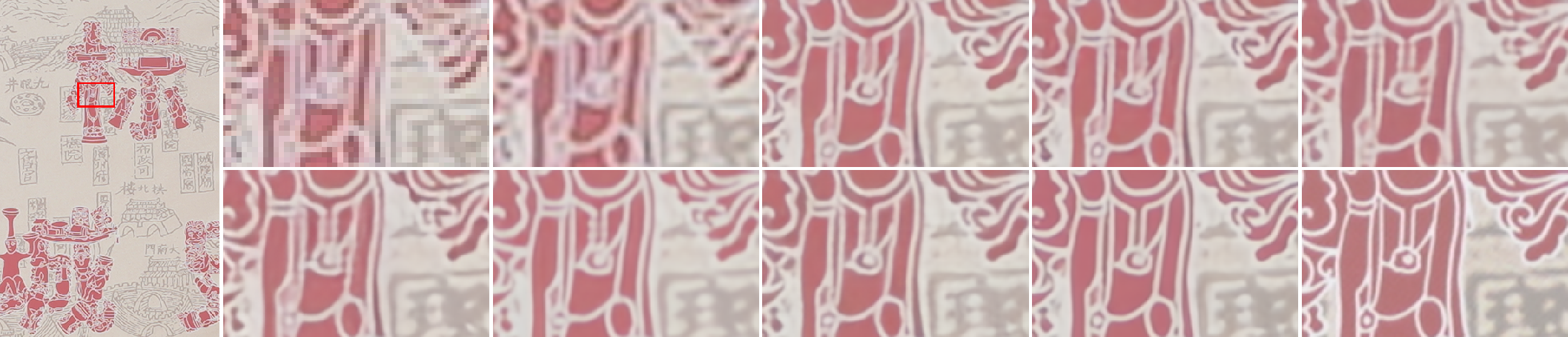}
  \includegraphics[width=\linewidth]{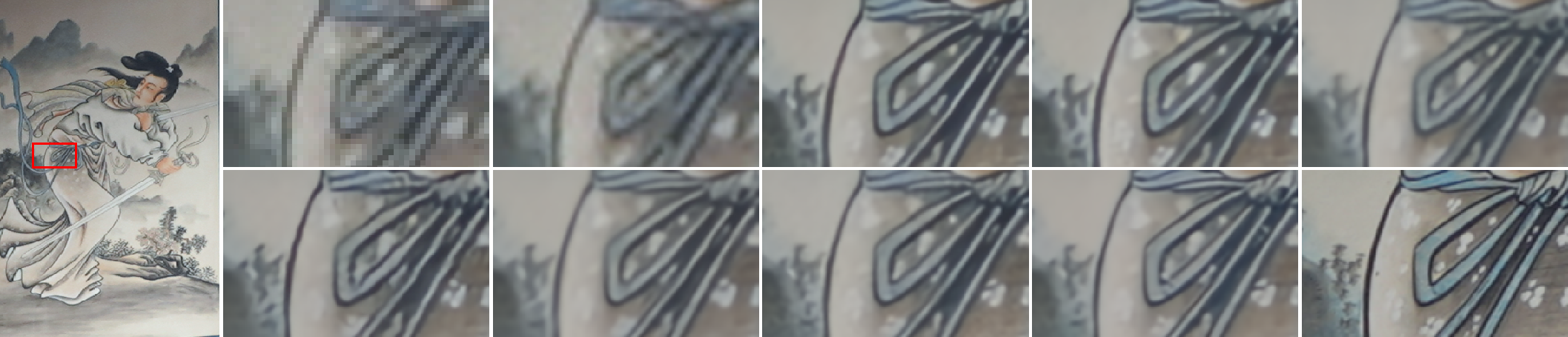}
  \caption{Qualitative comparison of different methods and our proposed BurstMamba on the RealBSR-RGB dataset for $\times4$ BISR with $L=14$.}
  \label{app:fig:sup_qualitative_result2}
\end{figure*}

\begin{figure*}[t]
  \centering
  \includegraphics[width=\linewidth]{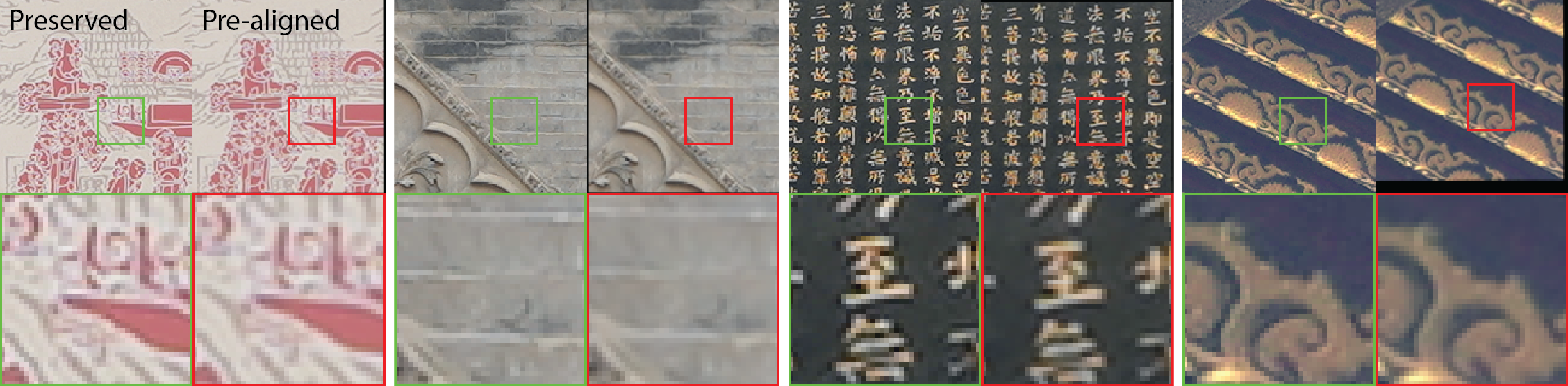}
  \caption{More examples to illustrate that pre-alignment of burst frames can impact the high frequency components in an image. For each image pairing, (left) original sample image from the burst sequence, (right) image that was aligned to the keyframe. We zoom in on identical regions across each image pair that contain high frequency components to demonstrate the loss of detail and sharpness in the aligned images.}
  \label{app:fig:dbs_visual2}
\end{figure*}

\begin{figure*}[t]
  \centering
  \includegraphics[width=\linewidth]{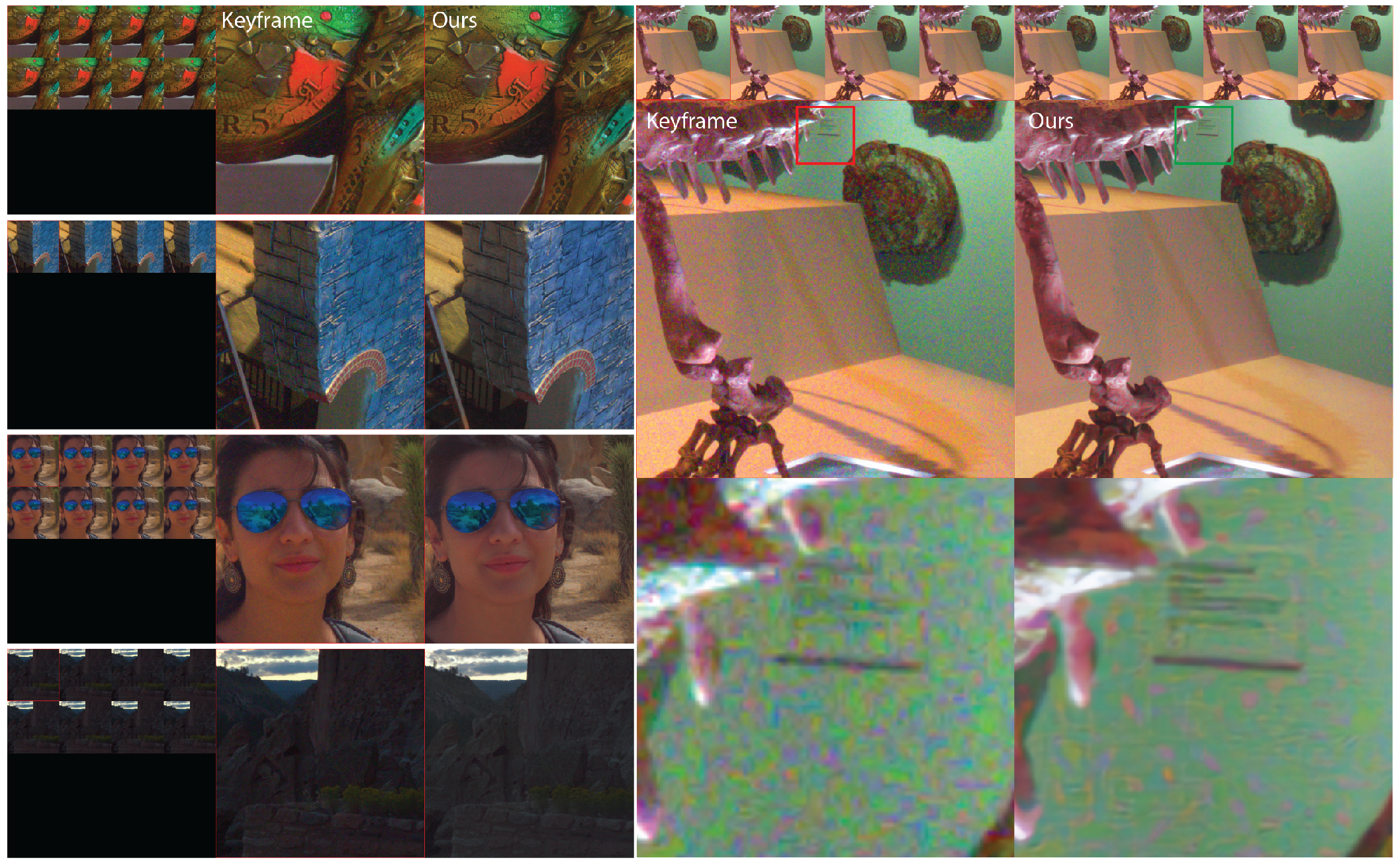}
  \caption{Qualitative out-of-distribution results of BurstMamba trained on RealBSR-RGB dataset with $L=14$ and evaluated on the HDR+ burst photography dataset with varying $L \in \{4,8\}$. By zooming into the example on the right, we again disclose and bring to light the limitations of BusrtMamba, having trained on a BISR dataset where high levels of noise are not a factor for reconstruction quality. While the text on the nameplate has much better high-frequency information, the emerging denoising capability of the model causes unwanted spot artifacts.}
  \label{app:fig:ood_qualitative_result2}
\end{figure*}

Additional qualitative results are provided in Fig.~\ref{app:fig:sup_qualitative_result1}, ~\ref{app:fig:sup_qualitative_result2}, more examples of loss of detail and sharpness in aligned images in Fig.~\ref{app:fig:dbs_visual2}, further out-of distribution samples from the HDR+ dataset in Fig.~\ref{app:fig:ood_qualitative_result2}, isolation of the burst length's effects in Fig.~\ref{app:fig:qualitative_ablation3}.

\begin{table}[t]
  \centering
  \caption{\label{app:tab:ablation_number_of_images_post}
  \newtext{Varying burst length at inference with a fixed L=14-trained model}}
  \begin{tabular}{|c|cc|}
    \hline
    $\#$ & PSNR$\uparrow$ & SSIM$\uparrow$ \\
    \hline
    1 & 31.289 & 0.890 \\
    2 & 32.171 & 0.907 \\
    5 & 32.764 & 0.918 \\
    10 & 33.156 & 0.926 \\
    14 & \textbf{33.287} & \textbf{0.929} \\
    \hline
  \end{tabular}
\end{table}

\begin{figure*}[t]
  \centering
  \comparisonheaders
  \includegraphics[width=\linewidth]{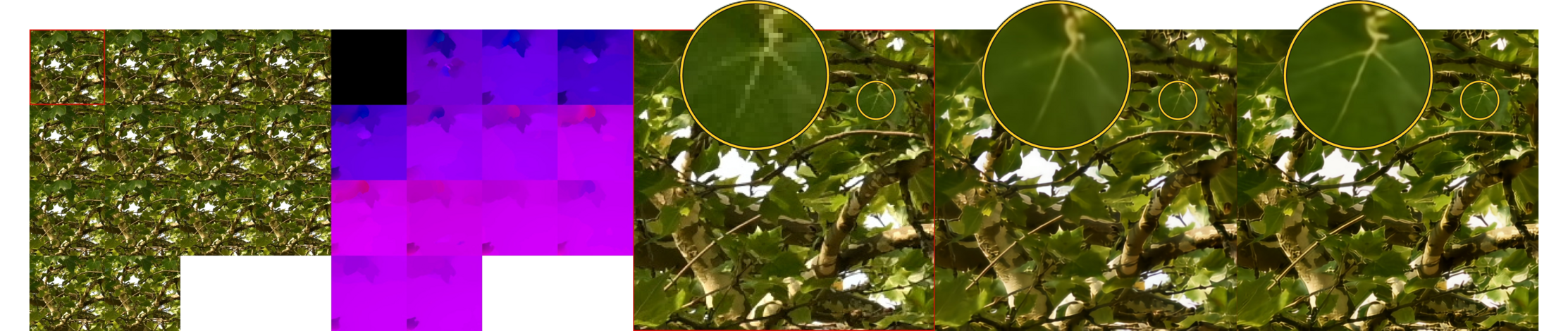}
  \includegraphics[width=\linewidth]{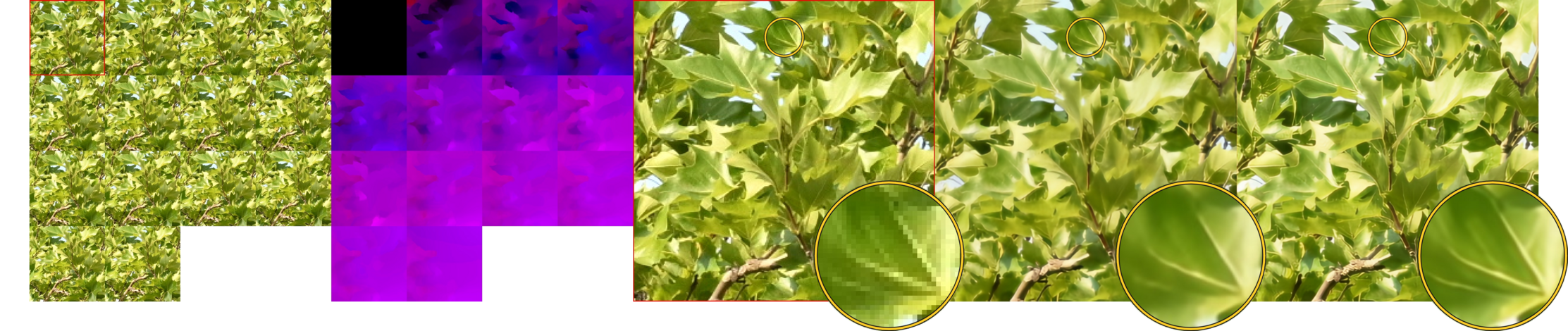}
  \includegraphics[width=\linewidth]{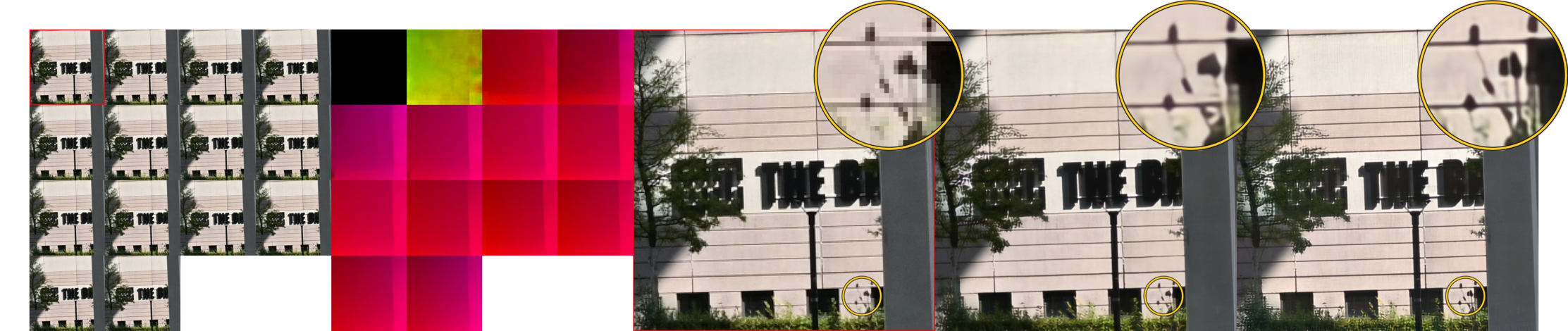}
  \caption{\newtext{Further OOD samples from our qualitative burst image dataset.}}
  \label{app:fig:our_ood1}
\end{figure*}

\begin{figure*}[t]
  \centering
  \makebox[\linewidth][c]{\begin{minipage}{1.05\linewidth}
    \comparisonheaders
    \includegraphics[width=\linewidth]{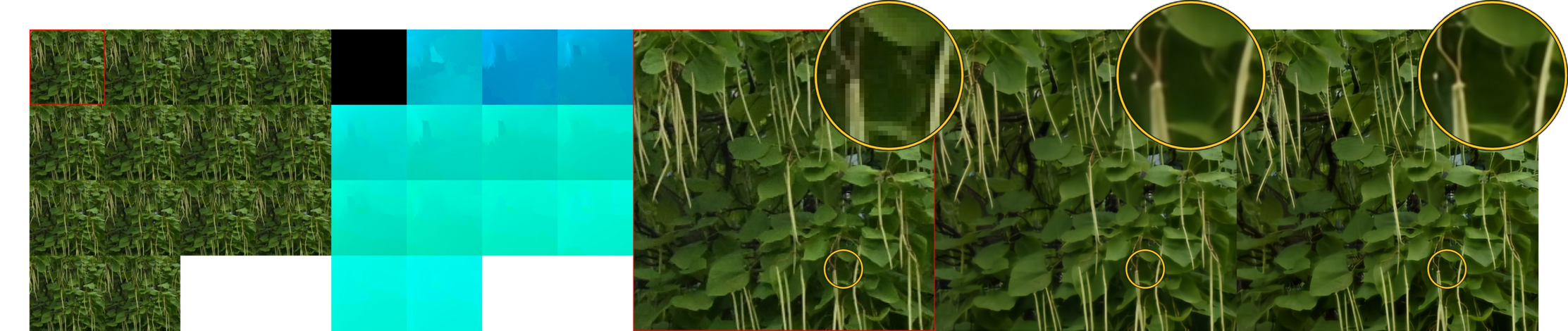}
    \includegraphics[width=\linewidth]{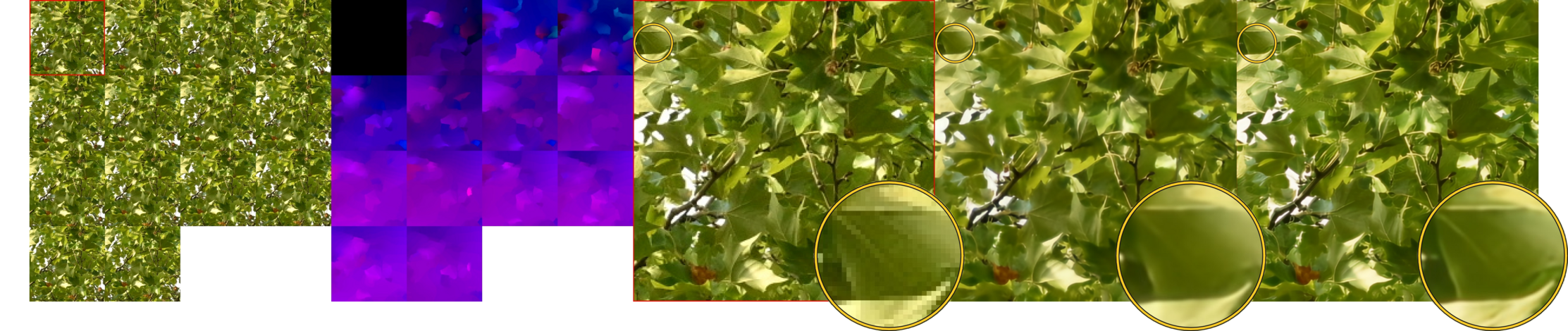}
    \includegraphics[width=\linewidth]{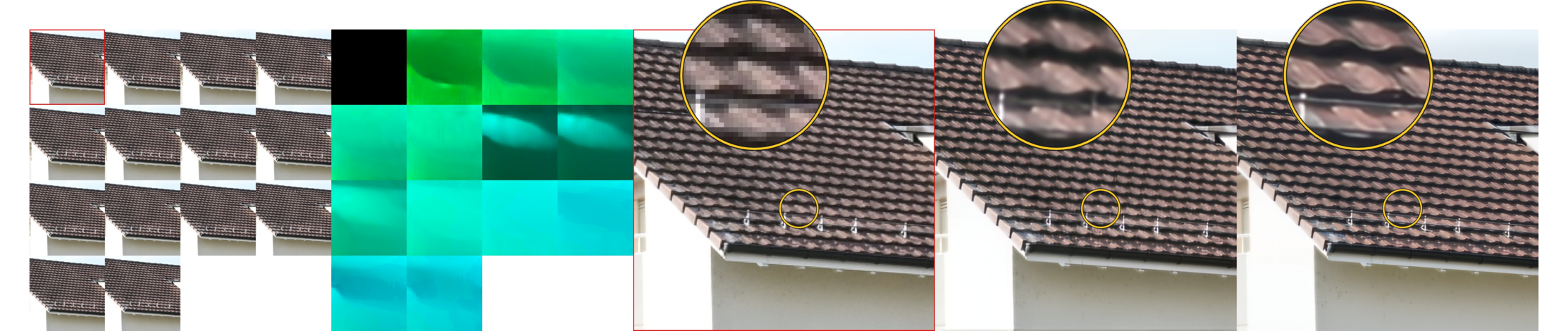}
    \includegraphics[width=\linewidth]{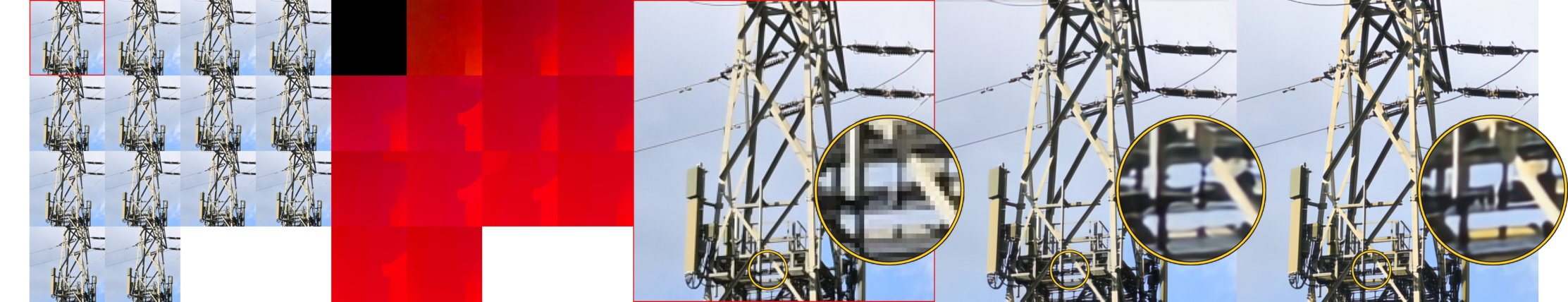}
  \end{minipage}}
  \caption{\newtext{Further OOD samples from our qualitative burst image dataset.}}
  \label{app:fig:our_ood2}
\end{figure*}

\begin{figure*}[t]
  \centering
  \makebox[\linewidth][c]{\begin{minipage}{1.05\linewidth}
    \comparisonheaders
    \includegraphics[width=\linewidth]{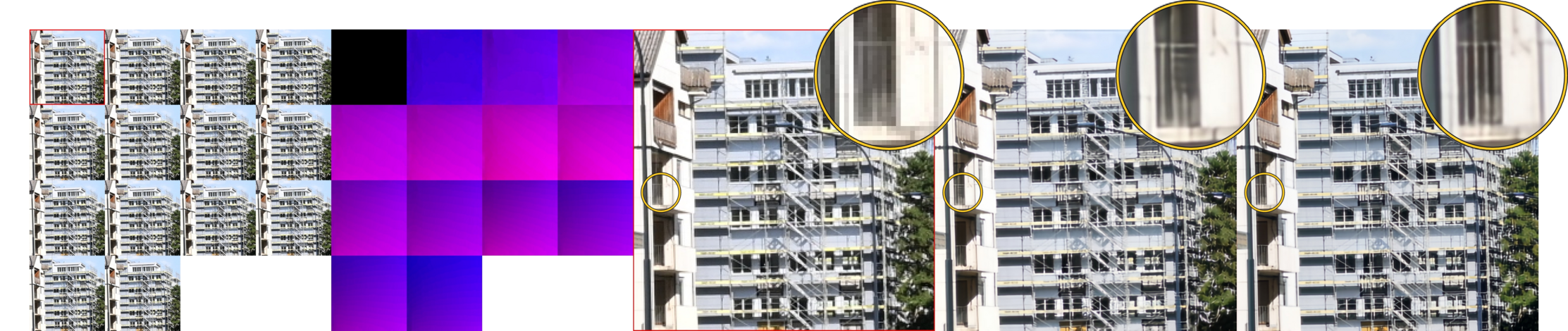}
    \includegraphics[width=\linewidth]{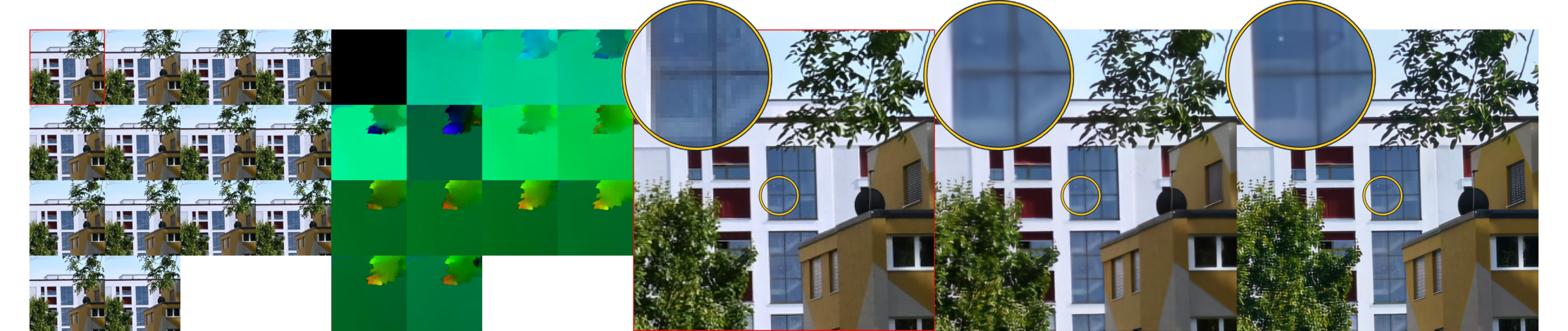}
    \includegraphics[width=\linewidth]{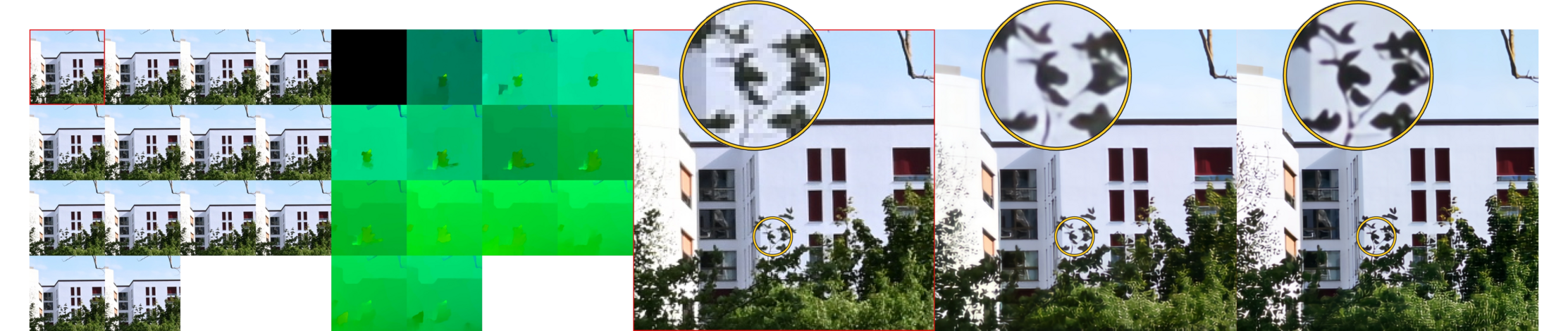}
    \includegraphics[width=\linewidth]{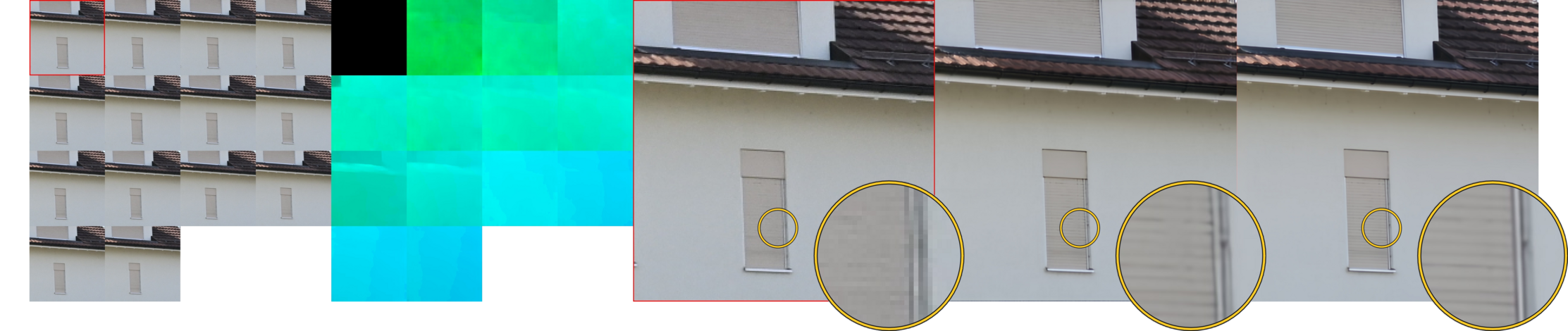}
  \end{minipage}}
  \caption{\newtext{Further OOD samples from our qualitative burst image dataset.}}
  \label{app:fig:our_ood3}
\end{figure*}

\paragraph{Qualitative Burst Image Dataset} \newtext{: We collected a qualitative out-of-distribution dataset using a Nikon Z9 equipped with a NIKKOR Z 24–70mm f/2.8 S zoom lens. The dataset is designed to evaluate publicly available models beyond the standard benchmark setting, with emphasis on challenging cases underrepresented in RealBSR-RGB, including vegetation, local motion, and fine repetitive structures. From the captured data, we selected 21 representative crop sequences with burst length L=14, enabling direct comparison with existing released checkpoints. All bursts were taken handheld with fixed exposure settings. We release the dataset at \url{ouenal.github.io/burstmamba/}.

Across the qualitative OOD examples in Fig.~\ref{app:fig:our_ood1}, Fig.~\ref{app:fig:our_ood2} and, Fig.~\ref{app:fig:our_ood3}, BurstMamba consistently recovers sharper and more coherent high-frequency detail than FBANet~\cite{wei2023towards}. This advantage is particularly visible on thin structures and repeated patterns, including leaf veins and stems, roof tiles, window frames, facade seams, blinds, and utility structures. FBANet generally produces plausible reconstructions but tends to oversmooth these regions, merging nearby edges and attenuating fine texture. In contrast, BurstMamba better preserves edge separation and structural continuity without introducing conspicuous ghosting in the highlighted regions. Since ground truth is unavailable for these scenes, these results provide qualitative evidence that BurstMamba generalizes more effectively to challenging real-world bursts with complex motion and previously unseen content.}

\subsection{Additional Ablations}
\label{app:sec:additional-ablations}

\begin{figure*}
  \centering
  \includegraphics[width=.98\linewidth]{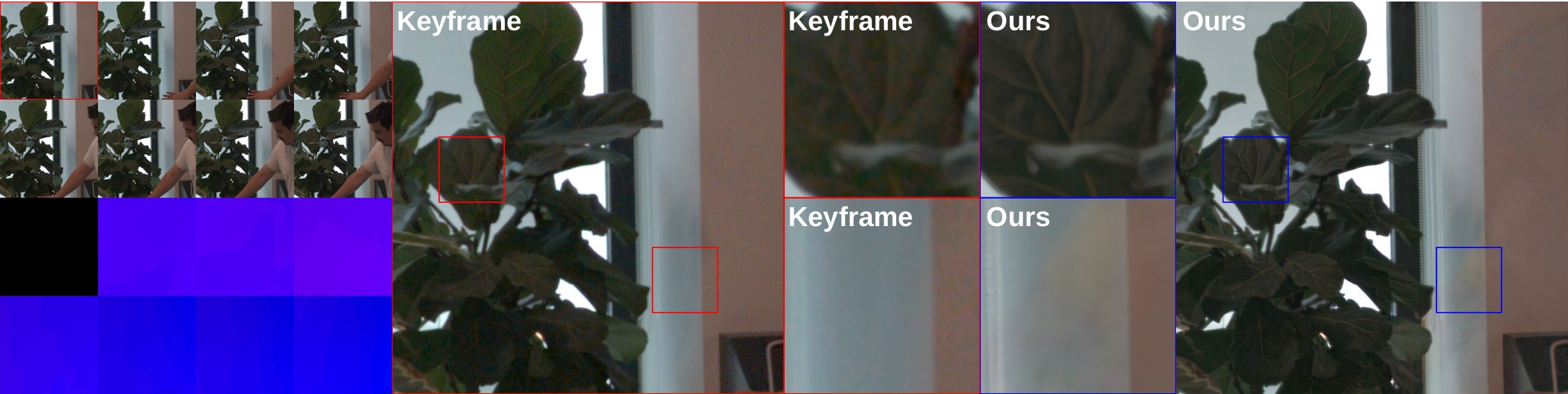}

  \caption{Out-of-distribution example with strong local motion and erroneous correspondence. While the background (e.g., the plant) is reconstructed faithfully, incorrect optical flow leads to mismatched aggregation in the moving regions, resulting in localized ghosting artifacts. This highlights a limitation under correspondence failures not represented in current BISR training data.}
  \label{app:fig:ood_qualitative_result3}
\end{figure*}

\paragraph{Robustness under optical flow failures.} In the main manuscript, we demonstrate the effectiveness of GAS using samples with moving salient objects, where the method reconstructs fine structures on and behind a moving robotic arm, despite not being explicitly trained for strong local motion. However, this lack of targeted training also exposes a limitation: when optical flow estimation fails, the model correspondingly degrades. In Fig.~\ref{app:fig:ood_qualitative_result3}, we present an additional OOD example where large motion is present but the estimated flow is incorrect. In this case, the model reconstructs the static background (e.g., the plant) well, but exhibits ghosting artifacts in the moving regions due to its reliance on erroneous correspondence cues. We argue that this limitation is not inherent to the formulation, but rather a consequence of the training distribution, and could be mitigated by incorporating examples with strong local motion and correspondence failures to improve robustness.

\begin{figure*}[t]
  \centering
  \includegraphics[width=1.017\linewidth]{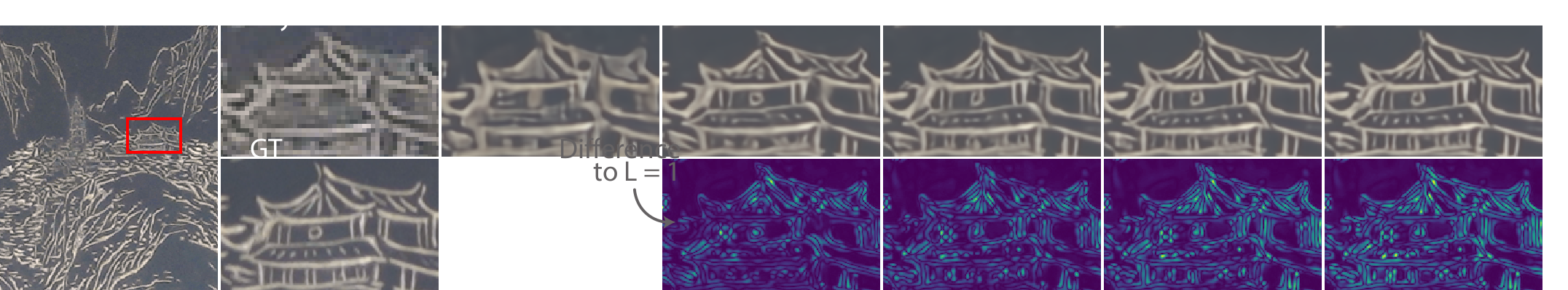}
  \includegraphics[width=\linewidth]{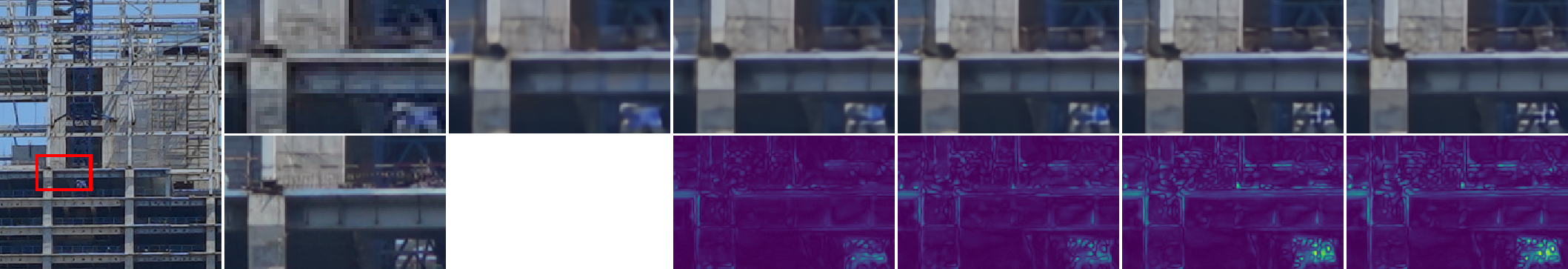}
  \caption{Further qualitative results from varying the input burst sequence length (L) for BurstMamba on the RealBSR-RGB dataset.}
  \label{app:fig:qualitative_ablation3}
\end{figure*}

\paragraph{Scaling with burst length and detachability.} \newtext{Tab.~\ref{app:tab:ablation_number_of_images_post} reports the performance of a single fixed trained model evaluated under varying burst lengths $L$ at inference, without any retraining. Performance increases monotonically as $L$ grows from 1 to 14, with diminishing returns for larger $L$.} Fig.~\ref{app:fig:qualitative_ablation3} provides additional qualitative examples across different burst lengths, complementing the analysis in the main paper.

\end{document}